\documentclass{article}

\usepackage{PRIMEarxiv}

\usepackage[utf8]{inputenc} % allow utf-8 input
\usepackage[T1]{fontenc}    % use 8-bit T1 fonts
\usepackage{hyperref}       % hyperlinks
\usepackage{url}            % simple URL typesetting
\usepackage{booktabs}       % professional-quality tables
\usepackage{amsfonts}       % blackboard math symbols
\usepackage{nicefrac}       % compact symbols for 1/2, etc.
\usepackage{microtype}      % microtypography
\usepackage{lipsum}
\usepackage{graphicx}

\usepackage{natbib}
\usepackage{amsthm}
\usepackage{graphicx} % Required for inserting images
\usepackage{booktabs} 
\usepackage{array}    
\usepackage{makecell} 
\usepackage{siunitx} 
\usepackage{amsmath}
\usepackage{subfigure}
\usepackage{multirow}
\usepackage[table]{xcolor} 

\newcolumntype{C}{>{\centering\arraybackslash}X}
\newcolumntype{R}{>{\raggedleft\arraybackslash}X}

\usepackage{amsfonts}
\usepackage{amsmath}
\usepackage{amssymb}
\usepackage{dsfont}
\usepackage{graphicx}
\usepackage{subfigure}
\usepackage{subcaption}
\usepackage{textcomp}
\usepackage{color}
\usepackage{adjustbox} %this
\usepackage{booktabs}
\usepackage{multirow}
\usepackage{soul}
\usepackage{comment}
\usepackage[table]{xcolor}
\usepackage{tikz}
\usepackage[linesnumbered,ruled,vlined]{algorithm2e}
\usetikzlibrary{matrix,positioning,arrows.meta}
\graphicspath{{media/}}     % organize your images and other figures under media/ folder

\title{Efficient Handwriting-Based Alzheimer’s Disease Diagnosis Using a Low-Rank Mixture of Experts Deep Learning Framework}

\author{
  Wu Wang, Yuang Cheng \\
  Center for Applied Statistics and School of Statistics \\
  Renmin University of China \\
  Beijing 100872, China \\
  \texttt{\{wu.wang, yuang.cheng\}@ruc.edu.cn} \\
  \And
  Fouzi Harrou, Ying Sun \\
  King Abdullah University of Science and Technology (KAUST) \\
  Computer, Electrical and Mathematical Sciences and Engineering (CEMSE) Division \\
  Thuwal 23955-6900, Saudi Arabia \\
  \texttt{\{fouzi.harrou, ying.sun\}@kaust.edu.sa} \\
}

\begin{document}
\maketitle

\begin{abstract}
Early and reliable detection of Alzheimer’s disease (AD) is crucial for timely clinical intervention and improved patient management. It also supports the evaluation of emerging therapeutic strategies. In this paper, we propose a Low-Rank Mixture of Experts (LoRA-MoE) deep learning framework for Alzheimer’s disease diagnosis based on handwriting analysis. Handwriting signals provide a non-invasive and scalable digital biomarker that captures subtle cognitive–motor impairments associated with early AD progression. The proposed architecture allows multiple experts to specialize in different handwriting patterns while sharing a common base network. This design enables efficient learning of general representations while reducing interference between experts. Each expert is equipped with lightweight low-rank adapters. This mechanism significantly reduces the number of trainable parameters compared with standard Mixture of Experts (MoE) models and improves training stability. The proposed framework is evaluated on the Diagnosis AlzheimeR WIth haNdwriting (DARWIN) dataset. Extensive experiments are conducted, including ablation studies on key architectural parameters such as hidden dimension size, number of experts, and LoRA rank. The method is compared with multilayer perceptron (MLP) and conventional MoE architectures. In addition, stacking ensemble strategies (StackMean and StackMax) are investigated to improve robustness and predictive performance. Experimental results show that the LoRA-MoE framework achieves powerful diagnostic performance while activating significantly fewer parameters during inference. These results highlight the potential of the proposed approach as an accurate and computationally efficient solution for handwriting-based Alzheimer’s disease screening and digital health applications.

\end{abstract}

% keywords can be removed
\keywords{Alzheimer's disease \and Handwriting \and deep learning \and unbalanced data \and stacked models.}

\section{introduction}
Alzheimer’s disease (AD) is a progressive neurodegenerative disorder and the leading cause of dementia worldwide, posing significant clinical, social, and economic challenges~\cite{jack2018nia,scheltens2021alzheimer}. Early and accurate diagnosis is critical for timely intervention, patient management, and the evaluation of emerging therapeutic strategies. However, current diagnostic procedures largely rely on invasive, expensive, or resource-intensive modalities such as neuroimaging, cerebrospinal fluid biomarkers, and extensive neuropsychological assessments, which limit their scalability for large-scale screening and continuous monitoring~\cite{blennow2015clinical,weiner2013alzheimer}.

\medskip
In recent years, machine learning (ML) and artificial intelligence (AI) have emerged as powerful tools for advancing early diagnosis across a wide range of healthcare applications, including neurodegenerative diseases such as Alzheimer’s disease~\cite{litjens2017survey}. Data-driven models have shown strong capability in extracting subtle and complex patterns from heterogeneous biomedical data, often outperforming traditional rule-based approaches~\cite{wang2024stacked}. For instance, in~\cite{wang2020early}, a deep learning framework that integrates premotor biomarkers achieved high accuracy for early detection of Parkinson’s disease. Manifold learning and semi-supervised anomaly detection have also been successfully applied to cardiovascular disease diagnosis using limited labeled data~\cite{harrou2025manifold}. In chronic disease monitoring, ML models leveraging thermal imaging have enabled accurate and deployable solutions for early diabetic foot ulcer detection~\cite{khandakar2021machine}. Beyond these domains, deep learning has demonstrated remarkable performance in cancer diagnosis from histopathology images~\cite{litjens2017survey} and in the automated detection of lung diseases from chest X-rays and CT scans~\cite{esteva2019guide}. Collectively, these studies highlight the potential of ML to support early, non-invasive, and scalable disease diagnosis, motivating its application to alternative biomarkers for Alzheimer’s disease prediction.

\medskip
 Recent studies have demonstrated that handwriting and speech signals constitute effective non-invasive digital biomarkers for detecting Alzheimer’s disease. Handwriting analysis captures subtle impairments in fine motor control, visuospatial organization, and executive planning, which are among the earliest cognitive functions affected by AD~\cite{neils2006dysgraphia,de2019handwriting}. Similarly, speech- and voice-based approaches exploit changes in fluency, lexical richness, prosody, and temporal dynamics to identify early cognitive decline using machine learning and deep learning models~\cite{luz2021alzheimer, fraser2015linguistic}. Compared to speech, handwriting offers several practical advantages: it is less sensitive to linguistic and cultural variability, easier to standardize across populations, and inherently reflects the interaction between cognitive and motor processes. These properties make handwriting particularly suitable for scalable and robust AD screening; however, effectively modeling the heterogeneity and high dimensionality of handwriting features remains a key challenge, motivating the advanced learning framework proposed in this study.

% SOTA methods

\medskip
In~\cite{kavitha2022early}, classical machine learning models, including Decision Tree, Random Forest, SVM, Gradient Boosting, and a Voting classifier, are applied to Alzheimer’s disease prediction using the OASIS neuroimaging dataset. The models are evaluated using accuracy, precision, recall, and F1-score, with ensemble methods achieving the best performance. The ensemble-based approach reports a maximum test accuracy of 83\%, demonstrating the effectiveness of traditional ML techniques for early AD diagnosis. In~\cite{kruthika2019multistage}, a multistage machine learning framework is proposed for Alzheimer’s disease diagnosis using MRI data from the ADNI database. The approach combines Naive Bayes, SVM, and KNN classifiers, with brain features extracted via FreeSurfer and optimized using particle swarm optimization (PSO). Experimental results show that the multistage strategy outperforms individual classifiers, highlighting the benefit of feature selection and classifier combination for early AD detection.

\medskip
Despite these promising advances, several limitations remain in existing machine learning approaches for Alzheimer’s disease diagnosis. First, many methods rely on high-dimensional biomedical modalities such as neuroimaging or speech signals, which may require specialized acquisition conditions or complex preprocessing. Second, conventional deep learning models, including multilayer perceptrons and standard mixture-of-experts architectures, often involve a large number of trainable parameters. This increases computational cost and may lead to overfitting when training data are limited, as is common in clinical studies. Third, handwriting-based diagnostic frameworks must capture heterogeneous cognitive–motor patterns across different writing tasks, which remains challenging for single-network models that lack explicit mechanisms for expert specialization. Consequently, there is a need for learning architectures that can efficiently model task variability while maintaining computational efficiency and robust generalization. These challenges motivate the development of the proposed Low-Rank Mixture of Experts (LoRA-MoE) framework, which combines expert specialization with parameter-efficient adaptation for handwriting-based Alzheimer’s disease diagnosis. The main contributions of this work are summarized as follows. 

\begin{itemize}
 \item An efficient deep learning architecture, termed Low-Rank Mixture of Experts (LoRA-MoE), is proposed for handwriting-based Alzheimer’s disease diagnosis. The architecture allows multiple experts to specialize in different handwriting patterns while reducing interference between tasks. This design enables the model to better capture heterogeneous cognitive–motor characteristics associated with AD.
\item A parameter-efficient expert learning strategy is introduced through low-rank adaptation. In the proposed LoRA-MoE framework, all experts share a common base network that learns general representations, while lightweight low-rank adapters enable task-specific specialization. Compared with standard MoE architectures, this design significantly reduces the number of trainable parameters, improves training stability, and mitigates catastrophic forgetting.
\item Extensive experiments are conducted on the DARWIN handwriting dataset to evaluate the proposed approach. The LoRA-MoE model is compared with a multilayer perceptron (MLP) and a conventional MoE architecture across different parameter configurations, including hidden dimension size, number of experts, and LoRA rank. The results demonstrate the effectiveness and robustness of the proposed framework for AD diagnosis.
\item Two stacking ensemble strategies, StackMean and StackMax, are investigated to further enhance classification performance. These strategies aggregate predictions from multiple model configurations, improving robustness and reducing the sensitivity to individual parameter settings.
\end{itemize}

%structure
\medskip
This paper is organized as follows. Section~\ref{sec2} introduces the proposed LoRA-MoE architecture, detailing its design and key components for efficient expert-based modeling. Section~\ref{sec3} presents the experimental results and discussion, with a specific focus on Alzheimer’s disease diagnosis using the DARWIN handwriting dataset. Finally, Section~\ref{sec4} concludes the paper by summarizing the main findings and highlighting the effectiveness of the proposed approach.

\section{The proposed LoRA-MoE Architecture}\label{sec2}

The concept of Mixture of Experts (MoE) was originally introduced in~\cite{jra1991moe,jra1993moe}. The fundamental idea of MoE is to decompose a complex learning task into several specialized submodels, referred to as experts, while a trainable gating network dynamically routes each input to a sparse subset of these experts. This mechanism allows different experts to specialize in distinct regions of the input space, thereby improving model capacity and efficiency. Recent advances in sparsely-gated MoE architectures~\cite{Shazeer2017OutrageouslyLN} have significantly expanded the applicability of this paradigm. In particular, their integration with transformer models and large-scale language models (LLMs)~\cite{Lepikhin2020GShardSG} has revitalized research in expert-based architectures and laid the foundation for modern MoE frameworks. Fedus et al.~\cite{fedus2022switch} proposed the Switch Transformer, demonstrating that a top-1 routing strategy, where each token is assigned to a single expert, can achieve competitive performance while drastically reducing computational overhead. More recently, Dai et al.~\cite{Dai2024DeepSeekMoETU} introduced DeepSeekMoE, where Fine-Grained Expert Segmentation and Shared Expert Isolation reduce parameter redundancy and enhance expert specialization. Similarly, Jiang et al.~\cite{Jiang2024MixtralOE} proposed Mixtral 8$\times$7B, which achieves performance surpassing traditional dense models while activating fewer than half of the parameters, illustrating the maturity and growing adoption of MoE architectures.

\medskip
Meanwhile, Low-Rank Adaptation (LoRA) was introduced in~\cite{hu2022lora} as a parameter-efficient fine-tuning technique for large neural networks. The core idea of LoRA is to approximate weight updates in dense layers by inserting trainable low-rank matrices, thereby significantly reducing the number of trainable parameters while maintaining performance comparable to full fine-tuning. Building on this concept, several extensions have been proposed, including ReLoRA~\cite{lialin2023relora}, an extendable and stackable fine-tuning strategy; AdaLoRA~\cite{zhang2023adalora}, which dynamically allocates rank based on singular value decomposition (SVD); and QLoRA~\cite{dt2023qlora}, which incorporates 4-bit quantization to further reduce memory requirements. Theoretical investigations have also confirmed LoRA's effectiveness. Malladi et al.~\cite{pmlr-v202-malladi23a} analyzed its fine-tuning dynamics from a kernel perspective and demonstrated that LoRA fine-tuning is nearly equivalent to full fine-tuning in the lazy training regime. Furthermore, Zeng et al.~\cite{zeng2024expressive} studied the expressive power of LoRA in fully connected and transformer networks, proving that under mild assumptions, LoRA can adapt a model to accurately represent any smaller target model.

\medskip
The modular, plug-and-play nature of LoRA enables it not only to be used independently but also to be combined through weighted aggregation. For example, LoRAHub~\cite{Huang2023LoraHubEC} automatically learns the optimal combination of LoRA modules trained on different tasks using a black-box optimization strategy, enabling cross-task generalization without requiring additional training. When LoRA modules remain trainable, jointly optimizing both the LoRA parameters and their aggregation weights can further enhance model performance. Following this idea, integrating LoRA with the MoE architecture~\cite{zadouri2023pushingmoe,feng2024moelora} represents a natural and effective extension. In such frameworks, each LoRA module acts as an expert, while a routing network determines each expert's contribution to the final prediction. This combination has demonstrated strong capabilities in various scenarios, including lifelong learning~\cite{yang2024moral}, mitigation of catastrophic forgetting~\cite{dou2024loramoe}, vision--language tasks~\cite{gou2024loramoevl}, and multi-task medical applications~\cite{liu2024loramoemed}.

\begin{figure}[h!]
\centering
\includegraphics[width=0.95\textwidth]{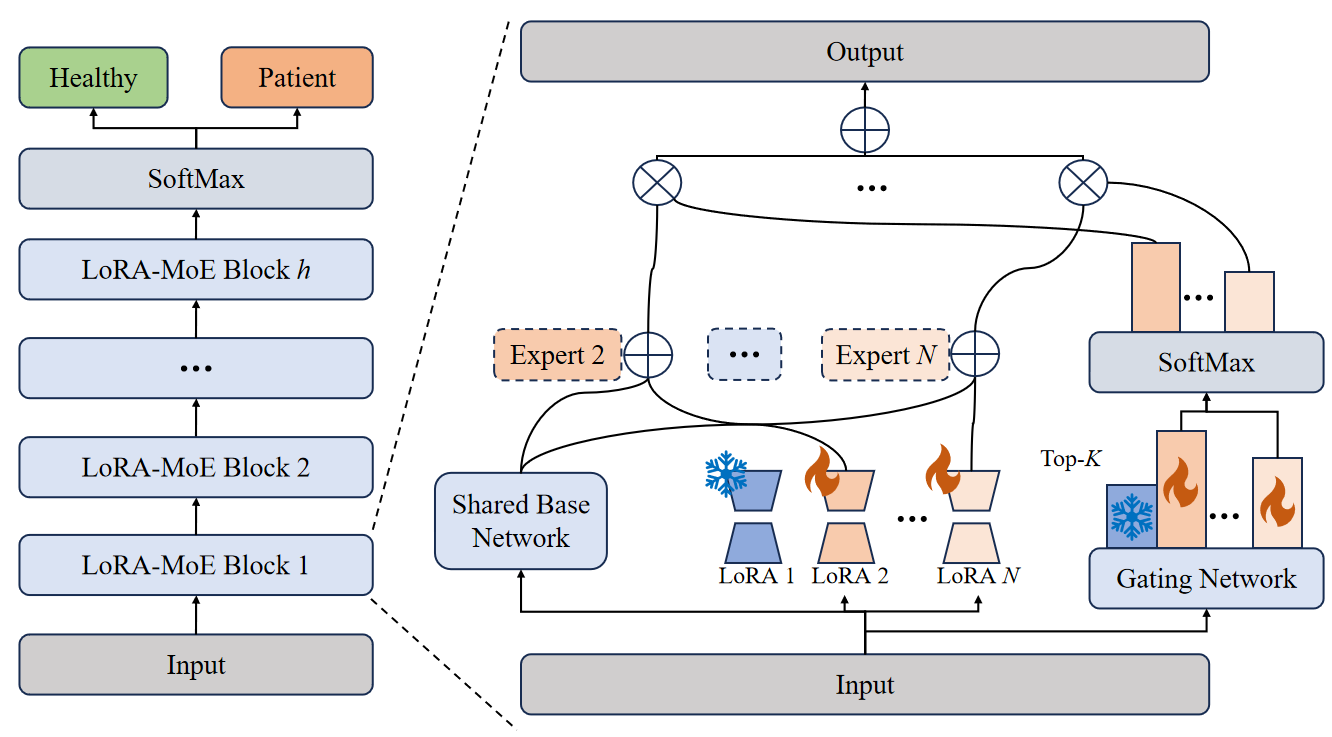}
\caption{Overview of the proposed LoRA-MoE architecture for handwriting-based Alzheimer’s disease diagnosis.}
\label{fig:flowchart}
\end{figure}

\medskip
In this context, we propose a LoRA-MoE architecture tailored for the diagnosis of Alzheimer’s disease (AD) from handwriting signals. The proposed model is designed to process feature inputs of varying lengths while maintaining parameter efficiency and strong representational capacity. Unlike conventional MoE models, in which each expert corresponds to a fully independent neural network, the proposed LoRA-MoE architecture decomposes experts into a shared base network and expert-specific LoRA adapters. This design significantly reduces parameter redundancy while enabling expert specialization. The LoRA-MoE architecture consists of three key components.  (1) The shared base network performs general feature extraction and learns representations shared across all experts.  (2) The gating network performs sparse routing and assigns each input sample to the most relevant experts.  (3) The LoRA adapters introduce lightweight low-rank adaptation modules that enable expert-specific specialization with minimal additional parameters. The final block produces a two-dimensional output corresponding to the class probabilities, which are converted into predictions using a softmax activation function. To enhance the representational capacity of the network, non-linear activation functions such as ReLU are employed between layers. For clarity of presentation, Figure~\ref{fig:flowchart} illustrates a single-layer LoRA-MoE structure.

% flowchart

\medskip
The main components of the proposed LoRA-MoE architecture are described in the following subsections. We first present the shared base network, followed by the gating network and the LoRA adapters. Finally, the prediction layer and the overall end-to-end workflow of the LoRA-MoE framework are described.

\subsection{Shared Base Network}
Shared Expert represents a significant innovation in MoE architectures in recent years. This design was first proposed in PR-MoE~\cite{rajbhandari2022deepspeed-moe}, where each token is processed by both a fixed expert and a selected expert through gating. DeepSeekMoE~\cite{Dai2024DeepSeekMoETU} adopted this design and achieved performance comparable to the state-of-the-art LLaMA2 model at the time, using only 40\% of its computational resources. It demonstrates the effectiveness of the shared expert mechanism.

In our proposed LoRA-MoE, the shared base network adopts a classical dense (fully connected) layer structure. Formally, let the input batch be denoted as $\mathbf{X} \in \mathbb{R}^{B \times d_{\text{in}}}$, where $B$ represents the batch size and $d_{\text{in}}$ denotes the input dimension. The output representations are denoted as $\mathbf{Y} \in \mathbb{R}^{B \times d_{\text{out}}}$, where $d_{\text{out}}$ is the output dimension. The shared base network is parameterized by a weight matrix $\mathbf{W} \in \mathbb{R}^{d_{\text{out}} \times d_{\text{in}}}$ and a bias vector $\mathbf{b} \in \mathbb{R}^{d_{\text{out}}}$. For a single input vector $\mathbf{x} \in \mathbb{R}^{d_{\text{in}}}$, the forward computation is defined as, 
\begin{equation}
\mathbf{y} = \mathbf{W}\mathbf{x} + \mathbf{b}.
\end{equation}

This structure forms the foundation of the entire model. Serving as a common feature-extraction backbone for all experts, it learns generic representations from the input handwriting features. These shared representations provide a stable foundation for subsequent expert-specific adaptations within the LoRA-MoE architecture. Unlike standard MoE models, in which each expert independently maintains a complete set of network parameters, all experts in LoRA-MoE share the base network's parameters. This design relies on two key assumptions. First, the tasks handled by different experts share a common underlying feature structure. Second, expert specialization primarily manifests in how these shared representations are adapted or weighted, rather than in the feature-extraction process itself. This strategy significantly reduces the total number of parameters while promoting knowledge sharing across experts.

\medskip
Let $N$ be the number of experts, $|\Theta_{\text{expert}}|$ denote the parameter count of a single expert network in standard MoE, and $|\Theta_{\text{lora}}|$ represent the parameter count of each LoRA adapter (i.e., the low-rank matrix pair). In standard MoE, the total parameter count is $N |\Theta_{\text{expert}}|$. In LoRA-MoE, the total parameter count equals that of the shared base network plus all LoRA adapters, i.e., $|\Theta_{\text{expert}}| + N |\Theta_{\text{lora}}|$. Because LoRA employs low-rank factorization, it generally holds that $|\Theta_{\text{lora}}| \ll |\Theta_{\text{expert}}|$. Consequently, the proportion of parameters that can be reduced by this method is approximated as: $$ 1 - \frac{|\Theta_{\text{expert}}| + N |\Theta_{\text{lora}}|}{N |\Theta_{\text{expert}}|} \approx 1 - \frac{1}{N} $$ This approximation is valid when $|\Theta_{\text{lora}}|$ is negligible relative to $|\Theta_{\text{expert}}|$.

As a concrete example, consider a layer with a bank of $N=5$ expert subnetworks, each having input and output dimensions of 300 ($d_{\text{in}}=d_{\text{out}}=300$). In a conventional standard MoE setup, each expert subnetwork contains $|\Theta_{\text{expert}}|=d_{\text{in}}\cdot d_{\text{out}}=90000$ parameters. For the LoRA-MoE variant, using low-rank adapters with rank $r=4$, the number of parameters per pair of adapter is $|\Theta_{\text{lora}}|=rd_{\text{in}}+rd_{\text{out}}=2400$. Under this configuration, the parameter reduction ratio achieved by replacing a conventional MoE layer with a LoRA-MoE layer can be expressed as $1 - \frac{|\Theta_{\text{expert}}| + N |\Theta_{\text{lora}}|}{N |\Theta_{\text{expert}}|}\approx77.3\%$. This illustrative example highlights the substantial gain in parameter efficiency achieved by the proposed design of the shared base network.

%Beyond substantially improving parameter efficiency to reduce storage and computational overhead, this design also helps mitigate overfitting through shared representations and enhances the stability of knowledge transfer among different experts.

Beyond improving parameter efficiency and reducing storage and computational overhead, this shared representation design also helps mitigate overfitting. By allowing experts to operate on a common representation space, it improves the stability of knowledge transfer among experts and enhances the generalization capability of the model when learning complex handwriting patterns associated with Alzheimer’s disease.

\subsection{Gating Network}
 The gating network adopts a shallow architecture to make routing decisions. It projects each input vector $\mathbf{x}$ onto a set of $N$ logits, where each logit corresponds to a candidate expert. In the simplest case, where the gating network consists of a single linear layer, the routing logits are computed as

\begin{equation}
\mathbf{g} = \mathbf{W}_g\mathbf{x} + \mathbf{b}_g, 
\quad 
\mathbf{W}_g\in \mathbb{R}^{N\times d_{\text{in}}},\;
\mathbf{b}_g \in \mathbb{R}^N
\end{equation}

\noindent where $\mathbf{g} \in \mathbb{R}^N$ represents the routing logits associated with the $N$ experts, and $N$ denotes the total number of experts. Sparse gating deploys experts through Top-$K$ strategy~\cite{Shazeer2017OutrageouslyLN}. That is, for each input sample, the gating network identifies the $K$ experts with the highest routing logits. A softmax operation is then applied only over this selected subset to produce aggregation weights $\tilde{\mathbf{g}}\in\mathbb{R}^K$. After computing the outputs from the selected experts, these weights are used to combine the expert outputs through a weighted sum. The final output for an input sample is therefore given by:
\begin{equation}
\mathbf{y} = \sum_{i=1}^{K} \tilde{g}_i \cdot \mathbf{E}_i(\mathbf{x})
\end{equation}
\noindent where $\mathbf{E}_i(\mathbf{x})$ denotes the output of the $i$-th selected expert. This selective routing mechanism enables the model to allocate computational resources to the most relevant expert subnetworks while maintaining overall efficiency.

\medskip 
The gating network directly receives standardized raw features rather than features extracted by deeper network layers. This design choice is motivated by an important consideration: routing decisions should remain independent of the feature transformations performed by the experts. Such independence avoids circular dependencies during training, where expert selection and feature extraction could influence each other, potentially leading to unstable optimization or convergence to suboptimal solutions.

For the Top-$K$ routing strategy, setting $K=1$ is often sufficient to achieve excellent performance~\cite{fedus2022switch,pmlr-v162-clark22a}. Top-1 routing means that only the expert with the highest routing logit is selected to process the current sample. This choice is motivated by three considerations.
\begin{enumerate}
\item Maximizing computational efficiency: During both training and inference, the proportion of activated parameters is $1/N$, which significantly reduces computational and memory overhead.

\item Promoting expert specialization: This routing strategy ensures that different experts receive largely non-overlapping samples, reinforcing a clear division of labor and encouraging feature decoupling among experts.

\item Ensuring decision consistency: Top-1 routing produces deterministic mappings, ensuring that similar inputs are consistently routed to the same expert as much as possible. This property supports reproducibility in applications such as medical diagnosis.
\end{enumerate}

This design maintains the expressive power of the model while achieving efficient conditional computation through sparse activation, and further facilitates the formation of stable and discriminative decision boundaries by individual experts. Specifically, consider the case where the gating network consists of a single linear layer. The Top-1 routing decision can be formalized as follows. Given an input $\mathbf{x} \in \mathbb{R}^d$, a weight matrix $\mathbf{W}_g \in \mathbb{R}^{N \times d}$, and a bias vector $\mathbf{b}_g \in \mathbb{R}^N$, the selected expert index is:
\begin{equation}
j^* = \arg\max_{j \in \{1,\dots,N\}} (\mathbf{w}_j^\top \mathbf{x} + b_j),
\end{equation}
\noindent where $\mathbf{w}_j^\top$ denotes the $j$-th row of $\mathbf{W}_g$. This decision rule is equivalent to maximizing a linear discriminant function.

Considering the decision boundary between any two experts $i$ and $j$, the boundary condition is
\begin{equation}
\mathbf{w}_i^\top \mathbf{x} + b_i
=
\mathbf{w}_j^\top \mathbf{x} + b_j,
\end{equation}

\noindent which can be rewritten as
\begin{equation}
(\mathbf{w}_i - \mathbf{w}_j)^\top \mathbf{x} + (b_i - b_j) = 0.
\end{equation}

This equation defines a hyperplane that divides the feature space $\mathbb{R}^d$ into two half-spaces,
\begin{itemize}
\item $H_{i>j} = \{\mathbf{x} \mid \mathbf{w}_i^\top \mathbf{x} + b_i > \mathbf{w}_j^\top \mathbf{x} + b_j\}$,
\item $H_{j>i} = \{\mathbf{x} \mid \mathbf{w}_j^\top \mathbf{x} + b_j > \mathbf{w}_i^\top \mathbf{x} + b_i\}$.
\end{itemize}

Considering the squared Euclidean distance:
\begin{equation}
\|\mathbf{x} + \tfrac{1}{2}\mathbf{w}_j\|^2
=
\mathbf{x}^\top \mathbf{x} + \mathbf{w}_j^\top \mathbf{x} + \tfrac{1}{4}\|\mathbf{w}_j\|^2,
\end{equation}

and comparing it with the linear function
\begin{equation}
\mathbf{w}_j^\top \mathbf{x} + b_j
=
\mathbf{w}_j^\top \mathbf{x}
+
\tfrac{1}{4}\|\mathbf{w}_j\|^2
+
\left(b_j - \tfrac{1}{4}\|\mathbf{w}_j\|^2\right),
\end{equation}
it follows that if $b_j - \tfrac{1}{4}\|\mathbf{w}_j\|^2$ is constant, maximizing $\mathbf{w}_j^\top \mathbf{x} + b_j$ is equivalent to minimizing $\|\mathbf{x} + \tfrac{1}{2}\mathbf{w}_j\|^2$. Ideally, the feature space can therefore be partitioned into $N$ regions
\begin{equation}
V(\mathbf{w}_j) =
\{\mathbf{x} \in \mathbb{R}^d
\mid
\|\mathbf{x} + \tfrac{1}{2}\mathbf{w}_j\|
\le
\|\mathbf{x} + \tfrac{1}{2}\mathbf{w}_i\|,
\;
\forall i \neq j
\}.
\end{equation}

This is a Voronoi partition with sites at $-\frac{1}{2}\mathbf{w}_j$, defining a nearest-neighbor partition: the cell $V(\mathbf{w}_j)$ is the set of all points whose distance to the site $\frac{1}{2}\mathbf{w}_j$ is not greater than their distance to any other site $\frac{1}{2}\mathbf{w}_i$.  Under the actual LoRA-MoE gating network, the weight vectors and bias terms do not necessarily satisfy the specific conditions. Nevertheless, the decision boundaries still retain key characteristics of a Voronoi partition, such as each input $\mathbf{x}$ being assigned to a unique expert and the acceptance region of each expert $R_j = \{\mathbf{x} \mid j = \arg\max_i f_i(\mathbf{x})\}$ forming a convex set. 

It demonstrates that, under Top-1 routing, the acceptance region of each expert corresponds to a convex polyhedron in the input feature space, and the decision boundaries collectively form a Voronoi-like partition based on linear discriminant functions. This structure provides a geometric interpretation of expert specialization: each expert is assigned a distinct, non‑overlapping region, ensuring deterministic routing. Such a partition not only promotes stable decision boundaries that enhance robustness but also aligns naturally with the design of LoRA‑MoE, where the shared base network defines a global refinement of the feature space while individual experts perform local, task‑specific refinements within their respective regions.

While Voronoi partition approximately achieved by Top-1 routing can naturally encourage experts to be specialized and it demonstrates good self-stability in our task, it may still encounter issues such as expert collapse or expert redundancy when applied to broader scenarios—where too few or too many samples are assigned to certain experts, causing their parameters to be updated either insufficiently or too frequently. To further enhance robustness, a load balancing loss~\cite{Lepikhin2020GShardSG} can be introduced as an explicit regularization term:
\begin{equation}
\mathcal{L}_{\text{balance}} = \lambda \cdot f(\mathbf{u}),
\end{equation}
\noindent where $\mathbf{u} = (u_1, \dots, u_N)$ is the vector of expert usage frequencies, $\lambda$ is the update rate, and $f$ is a function that measures data dispersion, such as the coefficient of variation or the Gini loss. A lower value of $f(\mathbf{u})$ indicates more balanced usage frequencies across all experts. This auxiliary loss is added as a separate term to the overall training objective and optimized jointly.

\subsection{LoRA Adapter}
The LoRA adapter layer operates in parallel with the shared base network and maintains a repository of $N$ pairs of low-rank matrices. Each LoRA adapter captures task-specific specialization, which is combined with the shared base network to form a specialized expert subnetwork. The core idea of LoRA is to enable efficient parameter updates during training through low-rank matrix factorization. In a conventional MoE architecture, each expert corresponds to a dense layer with parameter matrix $\mathbf{W}^{(j)}$. In LoRA-MoE, this matrix is replaced by the form $\mathbf{W} + \mathbf{A}^{(j)}\mathbf{B}^{(j)}$, where $\mathbf{W}$ denotes the shared base weight matrix, while $\mathbf{A}^{(j)}$ and $\mathbf{B}^{(j)}$ represent the low-rank adapter matrices specific to expert $j$.

Specifically, each expert $j$ is associated with the shared base network augmented by a pair of rank-$r$ matrices:
\begin{equation}
\mathbf{A}^{(j)} \in \mathbb{R}^{d_{\text{in}} \times r}, 
\quad
\mathbf{B}^{(j)} \in \mathbb{R}^{r \times d_{\text{out}}}.
\end{equation}

For the selected expert $j^*$, the LoRA adaptation is computed as:
\begin{equation}
\Delta^{(j^*)} = (\mathbf{x}\mathbf{A}^{(j^*)}) \mathbf{B}^{(j^*)} \cdot \left(\frac{\alpha}{r}\right),
\end{equation}
\noindent where $r $ is the rank of the low-rank matrices, $\alpha$ is a hyperparameter, and $\alpha/r$ acts as a scaling factor that controls the update rate of the parameters. The final output is obtained by combining the shared base output and the expert-specific adjustment term:
\begin{equation}
\mathbf{y} =
(\mathbf{W}\mathbf{x} + \mathbf{b}) + \Delta^{(j^*)}.
\end{equation}

The low-rank constraint $\text{rank}(\mathbf{A}^{(j)} \mathbf{B}^{(j)}) \leq r$ forces the expert's adjustment matrix $\Delta^{(j)}$ to reside in a low-dimensional subspace of at most $r$ dimensions. This design not only significantly reduces model complexity and effectively prevents overfitting, but also, owing to its parameter efficiency, makes the approach particularly suitable for scenarios with limited training data. Furthermore, it allows flexible expansion of total model capacity by increasing the number of experts $N$, while the shared base weight helps mitigate potential catastrophic forgetting often encountered in multi-expert training.

During practical training, the adapter matrices $\mathbf{A}$ and $\mathbf{B}$ are initialized using different strategies. The matrix $\mathbf{A}$ is initialized using the Kaiming uniform method~\cite{he2015ImageNet}, where its elements are independently sampled from the distribution
\begin{equation}
\mathcal{U}\left(-\sqrt{\frac{6}{d_{\text{in}}}}, \sqrt{\frac{6}{d_{\text{in}}}}\right).
\end{equation}
This initialization is designed for networks employing ReLU activation functions, as it stabilizes activation variance during forward propagation and preserves gradient magnitudes during backpropagation. In contrast, the matrix $\mathbf{B}$ is initialized to zero.

\medskip
Under this initialization scheme, the LoRA adjustment term $\Delta = (\mathbf{x}\mathbf{A})\mathbf{B}$ evaluates to zero at the beginning of training for any input $\mathbf{x}$. Consequently, the initial behavior of the LoRA-MoE model is entirely determined by the shared base network. During the first backpropagation step, the gradient of $\mathbf{B}$,
\begin{equation}
\frac{\partial \mathcal{L}}{\partial \mathbf{B}} =
\mathbf{A}^{\top}\left(\frac{\partial \mathcal{L}}{\partial \Delta}\right),
\end{equation}
becomes non-zero.  Because each expert is equipped with a unique $\mathbf{A}$, the gradients received by different $\mathbf{B}$ adapters point in diverse directions, thereby prompting each expert to evolve along specialized trajectories. This initialization strategy implies that training begins without expert-specific behavior and progressively learns differentiated adjustments. It avoids forcing experts to diverge along completely random directions at initialization, thereby improving training stability and promoting reliable convergence.

%\subsection{Prediction}

\subsection{Prediction and End-to-End Workflow of LoRA-MoE}
After the sequential processing through the $h-1$ LoRA-MoE blocks, the final block projects the high-dimensional feature sequence into a two-dimensional classification output. Then a softmax function is applied to produce a probability distribution, providing interpretable discrimination between 'Patient' and 'Healthy'. This classification step is critical for applying the extracted features into a diagnosis decision. The layer that compresses refined features into a two-dimensional output is still a LoRA-MoE, not a linear layer. This design ensures that the specialized representations learned by preceding experts can contribute differentially to the final classification via the gating mechanism, rather than being mixed and homogenized through a simple linear combination, rendering the specificity of the experts deactivated.

\begin{figure}[h!]
\centering
\includegraphics[width=0.95\textwidth]{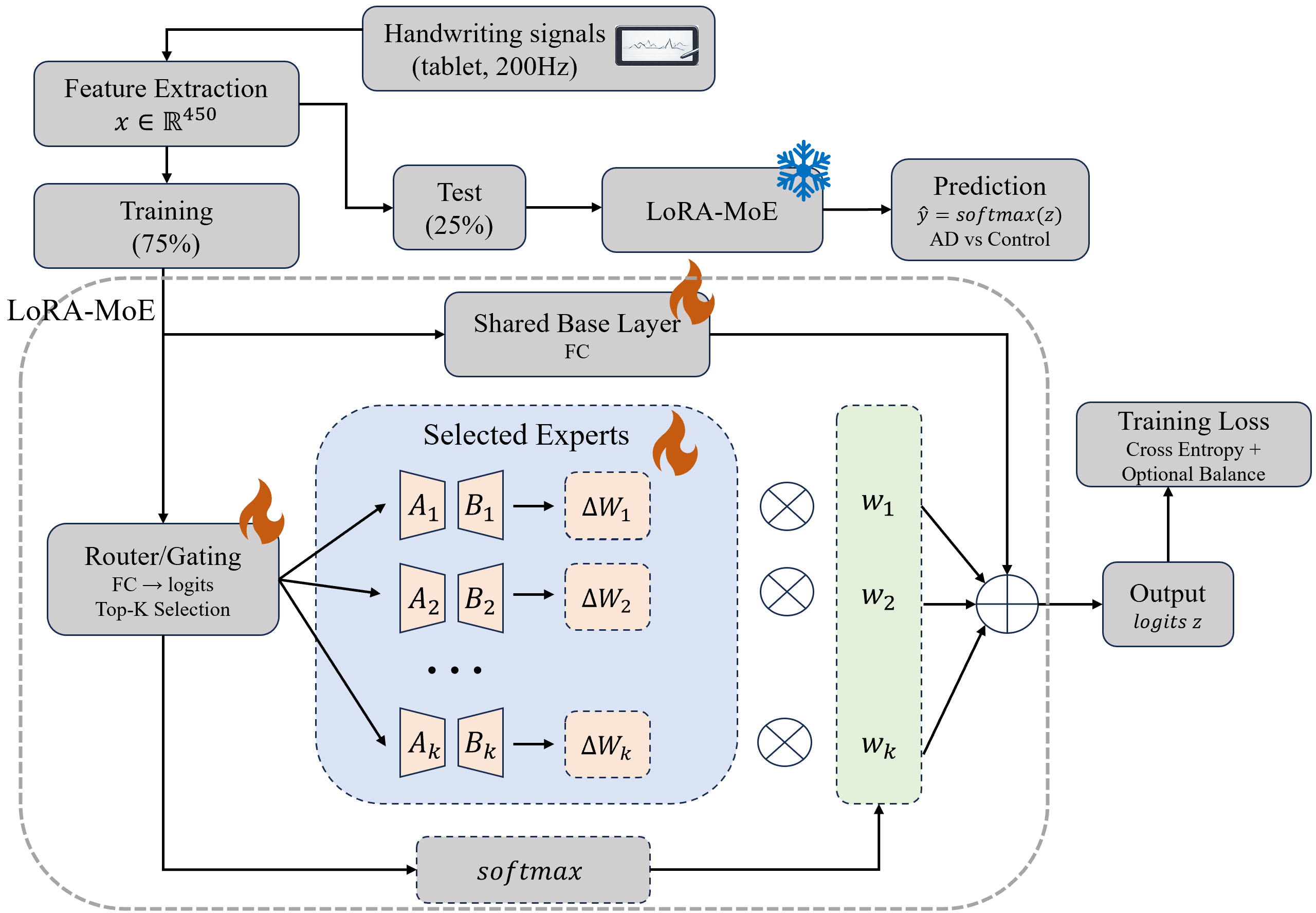}
\caption{The end-to-end workflow of the proposed LoRA-MoE architecture for Alzheimer’s disease diagnosis using handwriting dynamics}
\label{fig:endtoendflowchart}
\end{figure}

\medskip 
Figure~\ref{fig:endtoendflowchart} illustrates the end-to-end workflow of the proposed LoRA-MoE architecture for Alzheimer’s disease diagnosis using handwriting dynamics.

 The framework starts by acquiring handwriting signals using a digitizing tablet, capturing both on-paper and in-air movements. These signals reflect not only motor execution but also cognitive planning processes, both of which are known to be affected by neurodegenerative decline in Alzheimer’s disease. From the recorded signals, a set of clinically relevant features is extracted to represent temporal, kinematic, and pressure-related characteristics of handwriting. These features serve as the input to the learning architecture and provide a compact yet informative description of behavioral alterations associated with cognitive impairment. The shared base network first learns general representations of handwriting patterns common to both healthy individuals and AD patients. To address the heterogeneity of disease manifestations, the LoRA-MoE architecture employs a gating mechanism that dynamically selects the most relevant expert for each input sample. This enables the model to adapt its response to different handwriting styles and impairment patterns rather than relying on a single global decision function. Each activated expert is equipped with low-rank adaptation (LoRA) modules that refine the shared representation while maintaining parameter efficiency. These modules allow experts to specialize in capturing subtle handwriting alterations, such as tremor intensity, movement variability, and execution delays, features closely linked to cognitive decline, without significantly increasing model complexity. Finally, the outputs of the selected experts are combined to produce the classification decision (AD vs. control). By integrating shared learning with expert specialization and efficient adaptation, the LoRA-MoE framework provides a robust and scalable approach for handwriting-based Alzheimer’s disease diagnosis, enabling accurate detection while maintaining computational efficiency suitable for real-world screening applications.

\section{Results and discussion}\label{sec3}
This section presents the experimental evaluation of the proposed LoRA-MoE framework for handwriting-based Alzheimer’s disease diagnosis. The performance of the models is assessed using several evaluation metrics, including Accuracy, Sensitivity, Specificity, AUC, Precision, and F1-score. First, the DARWIN dataset used in this study is described. Next, an ablation analysis is conducted to investigate the influence of key architectural parameters, including the hidden dimension size, the number of experts, and the LoRA rank. We then evaluate the model on independent handwriting tasks and analyze the impact of model depth through multi-layer architectures. Finally, classification performance is examined across different handwriting tasks to assess the robustness and effectiveness of the proposed approach.

\subsection{Diagnosis AlzheimeR WIth haNdwriting (DARWIN) Dataset}

\begin{figure}[h!]
\centering
\includegraphics[width=0.7\linewidth]{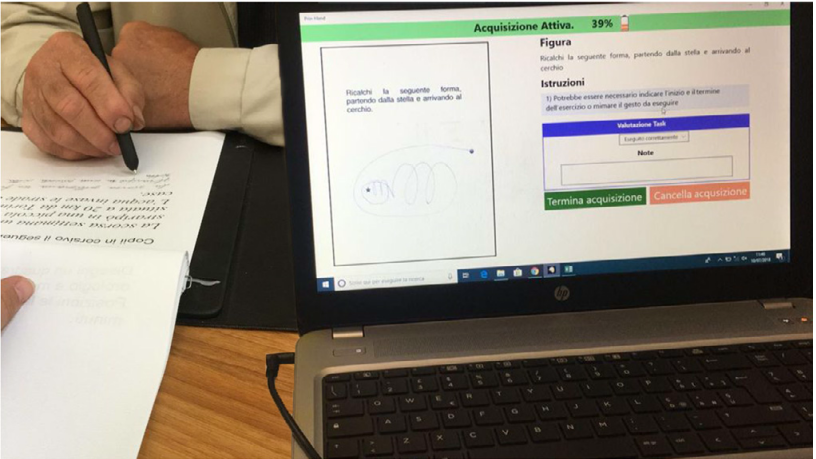}
\caption{Handwriting data acquisition setup using a graphic tablet to capture pen movements in real time~\cite{cilia2022diagnosing}.}
\label{fig:data_acquisition}
\end{figure}

The DARWIN dataset is a publicly available benchmark designed to support Alzheimer’s disease (AD) diagnosis through handwriting analysis. It provides clinically validated data collected from individuals diagnosed with AD and cognitively healthy controls, offering a non-invasive alternative to conventional diagnostic modalities. The dataset contains handwriting recordings from 174 participants, including 89 AD patients and 85 healthy controls. Clinical labels were assigned based on standard neuropsychological assessments, including the Mini-Mental State Examination (MMSE), Frontal Assessment Battery (FAB), and Montreal Cognitive Assessment (MoCA). Data acquisition followed a structured and standardized protocol to ensure consistency across participants and tasks. Handwriting signals were collected using a digitizing tablet equipped with a pressure-sensitive pen, allowing participants to write naturally on A4 paper placed on the tablet. The overall acquisition setup is illustrated in Figure~\ref{fig:data_acquisition}. For each task, the tablet recorded the pen-tip horizontal and vertical coordinates at a sampling frequency of 200~Hz. Two types of movements were captured: \emph{on-paper} movements, recorded when the pen tip was in contact with the paper together with the applied pressure, and \emph{in-air} movements, recorded when the pen was lifted from the paper within a short distance threshold. This distinction enables the analysis of both the execution and planning phases of handwriting, which are often affected by cognitive decline.

\medskip 
The acquisition protocol included 25 handwriting tasks designed to elicit diverse cognitive–motor behaviors. These tasks encompass basic graphic exercises targeting motor execution, copy and reverse-copy tasks assessing visuospatial planning and coordination, and memory- or dictation-based tasks evaluating recall and executive function. During data collection, participants were seated comfortably at a table, and the acquisition software displayed task instructions and real-time visual feedback, ensuring consistent task administration. From the raw handwriting signals, a set of handcrafted features was extracted to characterize cognitive and motor alterations associated with AD. These features are grouped into time-related, movement-related, and pressure-related categories, each capturing complementary aspects of handwriting behavior influenced by cognitive decline. A summary of the extracted features is provided in Table~\ref{tab:handwriting_features}. Time-related features reflect processing speed and motor efficiency, movement-related features describe kinematic properties and tremor, and pressure-related features quantify fine motor control. Together, these features form a compact yet expressive representation of handwriting dynamics suitable for learning-based AD diagnosis.
\begin{table}[h!]
\centering
\normalsize
\caption{Summary of features extracted from handwriting data.}
\label{tab:handwriting_features}
\begin{tabular}{ll}
\hline
\textbf{Category} & \textbf{Description} \\
\hline
\multicolumn{2}{l}{\textbf{Time-related features}} \\
\hline
total\_time & Total duration required to complete the task. \\
air\_time & Time spent performing in-air pen movements. \\
paper\_time & Time spent performing on-paper pen movements. \\

\hline
\multicolumn{2}{l}{\textbf{Movement-related features}} \\
\hline
mean\_speed\_on\_paper & Average speed of on-paper movements. \\
mean\_speed\_in\_air & Average speed of in-air movements. \\
mean\_acc\_on\_paper & Average acceleration of on-paper movements. \\
mean\_acc\_in\_air & Average acceleration of in-air movements. \\
mean\_jerk\_on\_paper & Average jerk of on-paper movements. \\
mean\_jerk\_in\_air & Average jerk of in-air movements. \\
gmrt\_on\_paper & Generalized mean relative tremor for on-paper movements. \\
gmrt\_in\_air & Generalized mean relative tremor for in-air movements. \\
mean\_gmrt & Average tremor across on-paper and in-air movements. \\
num\_of\_pendown & Number of pen-down events during writing. \\
max\_x\_extension & Maximum horizontal writing span. \\
max\_y\_extension & Maximum vertical writing span. \\
disp\_index & Spatial dispersion of the written trace. \\

\hline
\multicolumn{2}{l}{\textbf{Pressure-related features}} \\
\hline
pressure\_mean & Mean pressure exerted by the pen tip. \\
pressure\_var & Variability of pen pressure during writing. \\
\hline
\end{tabular}
\end{table}

\subsection{Ablation Study for Alzheimer’s Disease Diagnosis: Hidden Dimensions, Expert Count, and LoRA Rank}

\noindent \textbf{a) Ablation Study on Hidden Dimensions.}
This subsection analyzes the effect of hidden dimension size on Alzheimer’s disease diagnosis from handwriting data, comparing the proposed LoRA-MoE model with MoE and MLP baselines under identical settings. For fairness, the number of experts in LoRA-MoE and MoE is fixed to 6, while the hidden dimension varies from 50 to 400. The MLP baseline adopts a standard two-layer fully connected architecture: the first layer maps the 450-dimensional handwriting features to a hidden dimension, followed by a ReLU activation, and the second layer outputs two-class predictions. For example, with a hidden dimension of 200, the MLP contains approximately 90,000 parameters in the first layer and 202 parameters in the output layer. The dataset is split into 75\% for training and 25\% for testing, and performance metrics are averaged across the 25 handwriting tasks of the DARWIN dataset. The results reported in Table~\ref{tab:model_comparison_hidden_dim} show that LoRA-MoE consistently outperforms both MoE and MLP architectures across most hidden dimension settings.
\begin{table}[h!]
\centering
\caption{Model Comparison with Different Hidden Dimensions}
\label{tab:model_comparison_hidden_dim}
\small
\begin{tabular}{@{}llccccccc@{}}
\toprule
\textbf{Arch} & \textbf{Model} & \textbf{Accuracy} & \textbf{Sensitivity} & \textbf{Specificity} & \textbf{AUC} & \textbf{Precision} & \textbf{F1 Score} & \textbf{Time(s)} \\
\midrule

\multirow{10}{*}{\rotatebox[origin=c]{90}{LoRA-MoE}}
& BaseLearner\_1 & 0.8457 & 0.8500 & 0.8412 & 0.9239 & 0.8537 & 0.8446 & 0.69 \\
& BaseLearner\_2 & \textbf{0.8657} & \textbf{0.8667} & 0.8647 & 0.9250 & \textbf{0.8752} & \textbf{0.8647} & 0.67 \\
& BaseLearner\_3 & 0.8457 & 0.8500 & 0.8412 & 0.9237 & 0.8545 & 0.8447 & 0.70 \\
& BaseLearner\_4 & 0.8629 & 0.8556 & \textbf{0.8706} & 0.9221 & 0.8719 & 0.8617 & 0.72 \\
& BaseLearner\_5 & 0.8457 & 0.8611 & 0.8294 & 0.9314 & 0.8558 & 0.8443 & 0.74 \\
& BaseLearner\_6 & 0.8429 & 0.8444 & 0.8412 & 0.9255 & 0.8516 & 0.8417 & 0.96 \\
& BaseLearner\_7 & 0.8514 & 0.8556 & 0.8471 & 0.9185 & 0.8615 & 0.8502 & 1.00 \\
& BaseLearner\_8 & 0.8457 & 0.8556 & 0.8353 & 0.9173 & 0.8537 & 0.8447 & 1.02 \\
& StackMax & 0.8514 & 0.8611 & 0.8412 & 0.9376 & 0.8612 & 0.8502 & N/A \\
& StackMean & 0.8486 & 0.8556 & 0.8412 & \textbf{0.9379} & 0.8572 & 0.8475 & N/A \\
\midrule

\multirow{10}{*}{\rotatebox[origin=c]{90}{MoE}}
& BaseLearner\_1 & 0.7857 & 0.7722 & 0.8000 & 0.8964 & 0.7935 & 0.7844 & 0.74 \\
& BaseLearner\_2 & 0.8086 & 0.8056 & 0.8118 & 0.8935 & 0.8183 & 0.8073 & 0.72 \\
& BaseLearner\_3 & 0.8000 & 0.7722 & 0.8294 & 0.9039 & 0.8106 & 0.7990 & 0.74 \\
& BaseLearner\_4 & 0.8086 & 0.7944 & 0.8235 & 0.8928 & 0.8153 & 0.8075 & 0.83 \\
& BaseLearner\_5 & 0.8200 & 0.8167 & 0.8235 & 0.8964 & 0.8286 & 0.8190 & 0.87 \\
& BaseLearner\_6 & 0.8429 & 0.8611 & 0.8235 & 0.9108 & 0.8499 & 0.8416 & 1.13 \\
& BaseLearner\_7 & 0.8200 & 0.7944 & 0.8471 & 0.9052 & 0.8279 & 0.8191 & 1.22 \\
& BaseLearner\_8 & 0.8171 & 0.7889 & 0.8471 & 0.8962 & 0.8250 & 0.8162 & 1.24 \\
& StackMax & 0.8400 & 0.8222 & 0.8588 & 0.9307 & 0.8513 & 0.8387 & N/A \\
& StackMean & 0.8486 & 0.8333 & 0.8647 & 0.9369 & 0.8595 & 0.8474 & N/A \\
\midrule

\multirow{10}{*}{\rotatebox[origin=c]{90}{MLP}}
& BaseLearner\_1 & 0.8371 & 0.8444 & 0.8294 & 0.9216 & 0.8454 & 0.8360 & 0.20 \\
& BaseLearner\_2 & 0.8486 & 0.8556 & 0.8412 & 0.9172 & 0.8555 & 0.8476 & \textbf{0.18} \\
& BaseLearner\_3 & 0.8486 & 0.8556 & 0.8412 & 0.9168 & 0.8595 & 0.8471 & 0.19 \\
& BaseLearner\_4 & 0.8571 & 0.8611 & 0.8529 & 0.9222 & 0.8663 & 0.8560 & 0.20 \\
& BaseLearner\_5 & 0.8457 & 0.8611 & 0.8294 & 0.9193 & 0.8549 & 0.8443 & 0.20 \\
& BaseLearner\_6 & 0.8486 & 0.8556 & 0.8412 & 0.9221 & 0.8588 & 0.8471 & 0.26 \\
& BaseLearner\_7 & 0.8514 & 0.8556 & 0.8471 & 0.9230 & 0.8622 & 0.8501 & 0.28 \\
& BaseLearner\_8 & 0.8543 & 0.8611 & 0.8471 & 0.9216 & 0.8655 & 0.8529 & 0.28 \\
& StackMax & 0.8514 & 0.8556 & 0.8471 & 0.9230 & 0.8600 & 0.8503 & N/A \\
& StackMean & 0.8514 & 0.8556 & 0.8471 & 0.9225 & 0.8600 & 0.8503 & N/A \\
\bottomrule
\end{tabular}
\caption*{\textbf{Note:} BL\_1–BL\_8 correspond to models with hidden dimensions from 50 to 400 (step size 50). For LoRA-MoE and MoE, the number of experts is fixed to 6.}
\end{table}

\medskip
As shown in Table~\ref{tab:model_comparison_hidden_dim}, LoRA-MoE achieves the best overall performance, with a peak accuracy of 86.57\%, sensitivity of 86.67\%, and F1-score of 86.47\%, reflecting a strong balance between AD detection and false positive control. This peak occurs at an intermediate hidden dimension, indicating effective capacity utilization without overfitting. In contrast, the standard MoE exhibits lower and less stable performance across hidden dimensions and incurs a higher computational cost due to fully independent expert parameters. The MLP baseline achieves competitive accuracy with substantially lower runtime, but its performance saturates as model capacity increases, highlighting its limited ability to capture heterogeneous handwriting patterns. Finally, stacking strategies further improve robustness, particularly in terms of AUC, with LoRA-MoE reaching values close to 0.94, confirming the benefit of aggregating predictions across multiple configurations.

\begin{figure}[h!]
\centering
\includegraphics[width=\textwidth]{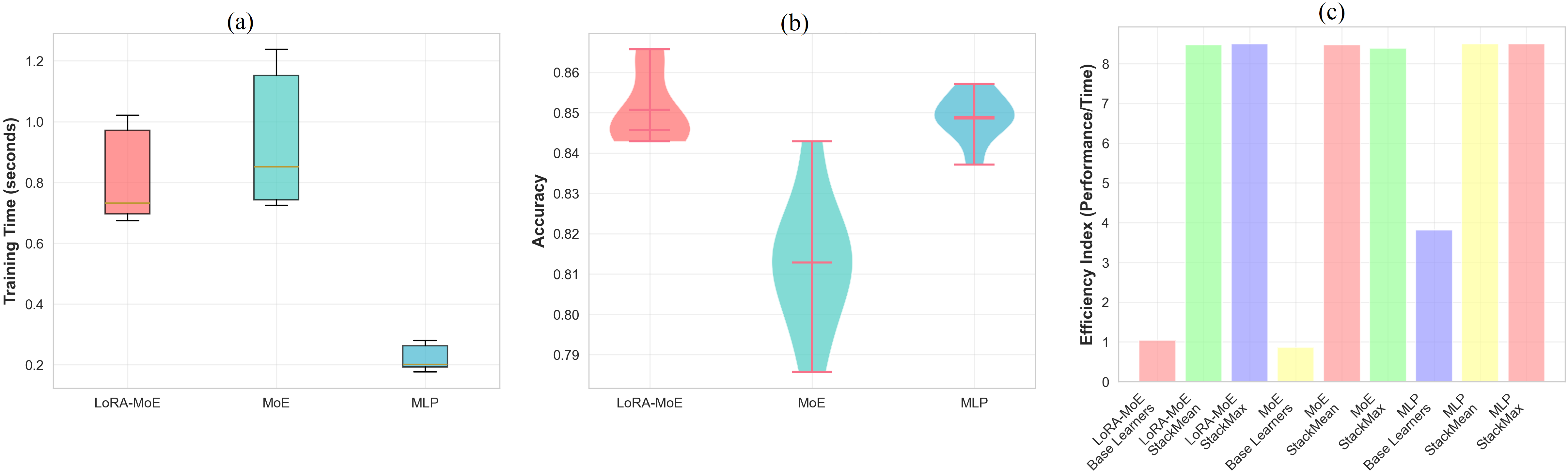}
\caption{Comparison of (a) training time, (b) base-learner accuracy distribution, and (c) efficiency index (performance/time) for LoRA-MoE, MoE, and MLP architectures.}
\label{fig:training_efficiency_analysis}
\end{figure}

\medskip
Fig.~\ref{fig:training_efficiency_analysis} visually summarizes training cost, performance stability, and efficiency across architectures. LoRA-MoE achieves lower and more stable training times than the standard MoE, highlighting the computational advantage of low-rank parameterization, while MLP is fastest but less expressive. The base-learner accuracy distributions show that LoRA-MoE maintains consistently high performance with limited variance, whereas MoE is more sensitive to architectural settings and MLP saturates at lower accuracy levels. The efficiency index further favors LoRA-MoE, particularly when combined with stacking strategies, confirming its superior trade-off between accuracy, robustness, and computational efficiency for handwriting-based Alzheimer’s disease diagnosis.

\noindent \textbf{b) Ablation Study on the Number of Experts.}
This subsection investigates the impact of varying the number of experts on Alzheimer’s disease diagnosis performance, with the objective of analyzing how expert diversity influences accuracy, robustness, and computational efficiency. The proposed LoRA-MoE model is compared against standard MoE and MLP baselines under identical settings. The results in Table~\ref{tab:model_comparison_n_expert} show that LoRA-MoE consistently benefits from increasing the number of experts up to an intermediate range, achieving its best performance with five experts.

\begin{table}[h!]
\centering
\caption{Model Comparison with Different Number of Experts}
\label{tab:model_comparison_n_expert}
\small
\begin{tabular}{@{}llccccccc@{}}
\toprule
\textbf{Arch} & \textbf{Model} & \textbf{Accuracy} & \textbf{Sensitivity} & \textbf{Specificity} & \textbf{AUC} & \textbf{Precision} & \textbf{F1 Score} & \textbf{Time(s)} \\
\midrule

\multirow{10}{*}{\rotatebox[origin=c]{90}{LoRA-MoE}}
& BL\_1 & 0.8514 & 0.8667 & 0.8353 & 0.9320 & 0.8545 & 0.8511 & 0.66 \\
& BL\_2 & 0.8514 & 0.8667 & 0.8353 & 0.9369 & 0.8538 & 0.8512 & 0.77 \\
& BL\_3 & 0.8543 & 0.8722 & 0.8353 & 0.9335 & 0.8590 & 0.8539 & 0.87 \\
& BL\_4 & 0.8686 & 0.8722 & \textbf{0.8647} & 0.9310 & 0.8714 & 0.8684 & 0.98 \\
& BL\_5 & \textbf{0.8714} & \textbf{0.8833} & 0.8588 & 0.9361 & \textbf{0.8751} & \textbf{0.8711} & 1.07 \\
& BL\_6 & 0.8486 & 0.8556 & 0.8412 & 0.9243 & 0.8525 & 0.8482 & 1.15 \\
& BL\_7 & 0.8486 & 0.8500 & 0.8471 & 0.9337 & 0.8527 & 0.8481 & 1.23 \\
& BL\_8 & 0.8657 & 0.8722 & 0.8588 & 0.9301 & 0.8689 & 0.8655 & 1.30 \\
& StackMax & 0.8657 & 0.8778 & 0.8529 & 0.9408 & 0.8688 & 0.8654 & N/A \\
& StackMean & 0.8657 & 0.8778 & 0.8529 & 0.9408 & 0.8686 & 0.8655 & N/A \\
\midrule

\multirow{10}{*}{\rotatebox[origin=c]{90}{MoE}}
& BL\_1 & 0.8400 & 0.8389 & 0.8412 & 0.9176 & 0.8422 & 0.8398 & 0.69 \\
& BL\_2 & 0.8457 & 0.8333 & 0.8588 & 0.9194 & 0.8519 & 0.8451 & 0.84 \\
& BL\_3 & 0.8429 & 0.8278 & 0.8588 & 0.9229 & 0.8506 & 0.8420 & 1.00 \\
& BL\_4 & 0.8286 & 0.8167 & 0.8412 & 0.9142 & 0.8334 & 0.8281 & 1.17 \\
& BL\_5 & 0.8114 & 0.8000 & 0.8235 & 0.9036 & 0.8245 & 0.8098 & 1.31 \\
& BL\_6 & 0.8171 & 0.8222 & 0.8118 & 0.9180 & 0.8205 & 0.8167 & 1.42 \\
& BL\_7 & 0.8057 & 0.7944 & 0.8176 & 0.9092 & 0.8153 & 0.8046 & 1.54 \\
& BL\_8 & 0.8057 & 0.7611 & 0.8529 & 0.8954 & 0.8158 & 0.8045 & 1.65 \\
& StackMax & 0.8571 & 0.8556 & 0.8588 & 0.9373 & 0.8619 & 0.8568 & N/A \\
& StackMean & 0.8629 & 0.8611 & \textbf{0.8647} & \textbf{0.9422} & 0.8680 & 0.8625 & N/A \\
\midrule

\multirow{10}{*}{\rotatebox[origin=c]{90}{MLP}}
& BL\_1 & 0.8600 & 0.8722 & 0.8471 & 0.9301 & 0.8629 & 0.8598 & \textbf{0.26} \\
& BL\_2 & 0.8571 & 0.8667 & 0.8471 & 0.9302 & 0.8600 & 0.8569 & 0.27 \\
& BL\_3 & 0.8600 & 0.8667 & 0.8529 & 0.9322 & 0.8630 & 0.8597 & \textbf{0.26} \\
& BL\_4 & 0.8571 & 0.8611 & 0.8529 & 0.9286 & 0.8604 & 0.8569 & \textbf{0.26} \\
& BL\_5 & 0.8600 & 0.8667 & 0.8529 & 0.9309 & 0.8628 & 0.8598 & 0.27 \\
& BL\_6 & 0.8543 & 0.8611 & 0.8471 & 0.9324 & 0.8570 & 0.8540 & 0.27 \\
& BL\_7 & 0.8543 & 0.8667 & 0.8412 & 0.9309 & 0.8579 & 0.8540 & \textbf{0.26} \\
& BL\_8 & 0.8571 & 0.8611 & 0.8529 & 0.9289 & 0.8598 & 0.8569 & 0.27 \\
& StackMax & 0.8600 & 0.8667 & 0.8529 & 0.9333 & 0.8630 & 0.8597 & N/A \\
& StackMean & 0.8600 & 0.8667 & 0.8529 & 0.9325 & 0.8630 & 0.8597 & N/A \\
\bottomrule
\end{tabular}
\caption*{\textbf{Note:} For LoRA-MoE and MoE, BL\_1 to BL\_8 correspond to the model with 3-10 experts, step is 1.For all models, their hidden dimension is equivalent to 300. For MLP, all BLs have the same structure and play a role in displaying random error.}
\end{table}

  At this configuration, LoRA-MoE reaches the highest accuracy (87.14\%), sensitivity (88.33\%), and F1-score (87.11\%), indicating improved detection of AD patients while maintaining balanced specificity. This trend highlights that a moderate number of specialized experts enables the model to effectively capture heterogeneous handwriting patterns without introducing redundancy or overfitting. In contrast, the standard MoE model exhibits less stable behavior as the number of experts increases. While performance improves slightly for a small number of experts, adding more experts leads to performance degradation and increased computational cost. This behavior reflects the limitations of conventional MoE architectures, where fully independent expert parameters increase optimization difficulty and overfitting susceptibility, particularly in small clinical datasets. The MLP baseline shows relatively stable performance across configurations, as its architecture remains unchanged regardless of the number of base learners. Although MLP achieves competitive accuracy with minimal computational cost, it lacks explicit expert specialization and therefore fails to benefit from increased model diversity. As a result, its performance saturates and does not scale with architectural complexity. Finally, stacking strategies further enhance robustness for both LoRA-MoE and MoE by aggregating complementary expert predictions. However, LoRA-MoE consistently achieves superior performance-efficiency trade-offs, combining high accuracy with controlled training time. These findings demonstrate that the proposed LoRA-MoE framework effectively leverages expert diversity through low-rank adaptation, making it more scalable and better suited for handwriting-based Alzheimer’s disease diagnosis than conventional MoE or single-network baselines.

\begin{figure}[h!]
\centering
\includegraphics[width=\textwidth]{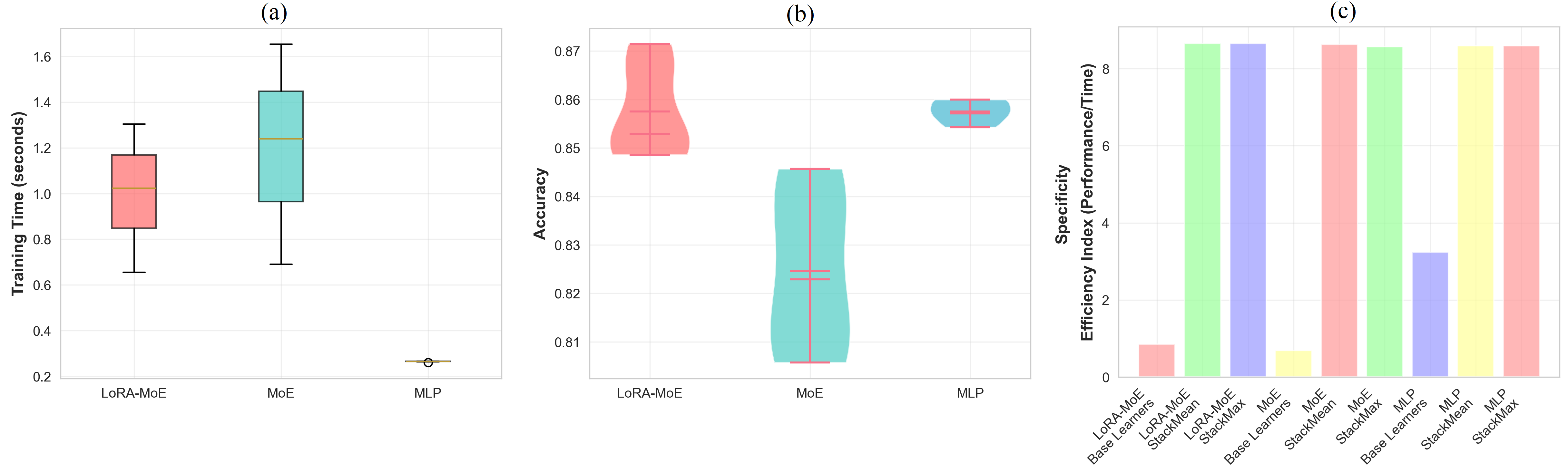}
\caption{Comparison of (a) training time, (b) base-learner accuracy distribution, and (c) efficiency index (performance/time) for LoRA-MoE, MoE, and MLP under different numbers of experts.}
\label{fig:expert_efficiency_analysis}
\end{figure}

\medskip
To further analyze the effect of expert cardinality, Fig.~\ref{fig:expert_efficiency_analysis} presents a visual comparison of training time, base-learner performance distribution, and efficiency across architectures with different numbers of experts. 

As shown in Fig.~\ref{fig:expert_efficiency_analysis}(a), LoRA-MoE maintains substantially lower and more stable training times than the standard MoE as the number of experts increases, confirming the computational advantage of sharing a common backbone with low-rank expert adaptation. In contrast, MoE exhibits a marked increase in training time, reflecting the overhead of fully independent expert parameters. The base-learner accuracy distributions in Fig.~\ref{fig:expert_efficiency_analysis}(b) further highlight the stability of LoRA-MoE. Its accuracy remains consistently high with limited variance across expert configurations, whereas MoE shows a wider dispersion, indicating sensitivity to the number of experts and less reliable optimization. The MLP baseline exhibits narrow variability but saturates at a lower performance level, as it lacks explicit expert specialization. Finally, the efficiency index shown in Fig.~\ref{fig:expert_efficiency_analysis}(c), defined as the ratio between performance and training time, clearly favors LoRA-MoE, particularly when combined with stacking strategies. Both StackMean and StackMax significantly improve efficiency by aggregating complementary expert predictions without incurring additional training cost. Overall, this analysis confirms that LoRA-MoE scales more effectively with expert diversity, offering a superior balance between accuracy, robustness, and computational efficiency for handwriting-based Alzheimer’s disease diagnosis.

\noindent \textbf{c) Ablation Study on Different Ranks of LoRA.}
This subsection evaluates the impact of the LoRA rank on model performance to identify an optimal trade-off between adaptation capacity and computational efficiency. The LoRA-MoE model is evaluated with ranks ranging from 1 to 8, while the hidden dimension and number of experts are fixed to 300 and 6, respectively. For comparison, MoE and MLP baselines retain identical architectures across all configurations, and variations in their results reflect random initialization effects rather than structural changes. Performance metrics are averaged across the handwriting tasks of the DARWIN dataset. Table~\ref{tab:model_comparison_rank} shows that LoRA-MoE performs best at low ranks, with rank~2 achieving the highest accuracy (85.14\%) and F1-score (85.11\%). Increasing the rank beyond this point yields no additional performance gains and incurs higher training costs, indicating that low-rank adaptation is sufficient to capture expert-specific handwriting patterns while avoiding over-parameterization. In contrast, MoE and MLP exhibit more stable but lower performance. Although stacking improves MoE robustness, LoRA-MoE consistently provides a superior balance between accuracy and computational efficiency for handwriting-based Alzheimer’s disease diagnosis.
\begin{table}[h!]
\centering
\caption{Model Comparison with Different Rank of LoRA}
\label{tab:model_comparison_rank}
\small
\begin{tabular}{@{}llccccccc@{}}
\toprule
\textbf{Arch} & \textbf{Model} & \textbf{Accuracy} & \textbf{Sensitivity} & \textbf{Specificity} & \textbf{AUC} & \textbf{Precision} & \textbf{F1 Score} & \textbf{Time(s)} \\
\midrule

\multirow{10}{*}{\rotatebox[origin=c]{90}{LoRA-MoE}}
& BL\_1 & 0.8086 & 0.7722 & 0.8471 & 0.9160 & 0.8187 & 0.8075 & 0.97 \\
& BL\_2 & \textbf{0.8514} & 0.8389 & 0.8647 & 0.9252 & 0.8561 & \textbf{0.8511} & 1.09 \\
& BL\_3 & 0.8486 & 0.8278 & \textbf{0.8706} & 0.9257 & 0.8551 & 0.8480 & 1.10 \\
& BL\_4 & 0.8314 & 0.7944 & \textbf{0.8706} & 0.9240 & 0.8398 & 0.8307 & 1.14 \\
& BL\_5 & 0.8229 & 0.8000 & 0.8471 & 0.9224 & 0.8287 & 0.8223 & 1.16 \\
& BL\_6 & 0.8371 & 0.8111 & 0.8647 & 0.9224 & 0.8424 & 0.8368 & 1.16 \\
& BL\_7 & 0.8314 & 0.8167 & 0.8471 & 0.9234 & 0.8364 & 0.8310 & 1.21 \\
& BL\_8 & 0.8257 & 0.8056 & 0.8471 & 0.9217 & 0.8309 & 0.8254 & 1.22 \\
& StackMax & 0.8257 & 0.7944 & 0.8588 & 0.9245 & 0.8334 & 0.8251 & N/A \\
& StackMean & 0.8343 & 0.8111 & 0.8588 & 0.9245 & 0.8401 & 0.8339 & N/A \\
\midrule

\multirow{10}{*}{\rotatebox[origin=c]{90}{MoE}}
& BL\_1 & 0.8229 & 0.7778 & 0.8706 & 0.9026 & 0.8332 & 0.8218 & 1.19 \\
& BL\_2 & 0.8371 & 0.8056 & 0.8706 & 0.9183 & 0.8429 & 0.8368 & 1.18 \\
& BL\_3 & 0.8143 & 0.7556 & 0.8765 & 0.9013 & 0.8241 & 0.8135 & 1.19 \\
& BL\_4 & 0.8343 & 0.7944 & 0.8765 & 0.9098 & 0.8468 & 0.8331 & 1.19 \\
& BL\_5 & 0.8143 & 0.7611 & 0.8706 & 0.9126 & 0.8227 & 0.8133 & 1.17 \\
& BL\_6 & 0.8229 & 0.7944 & 0.8529 & 0.9082 & 0.8280 & 0.8225 & 1.16 \\
& BL\_7 & 0.8314 & 0.7889 & 0.8765 & 0.9098 & 0.8374 & 0.8310 & 1.15 \\
& BL\_8 & 0.8229 & 0.7833 & 0.8647 & 0.9029 & 0.8294 & 0.8222 & 1.14 \\
& StackMax & 0.8629 & 0.8167 & \textbf{0.9118} & 0.9271 & \textbf{0.8706} & 0.8624 & N/A \\
& StackMean & 0.8543 & 0.8056 & 0.9059 & \textbf{0.9337} & 0.8630 & 0.8537 & N/A \\
\midrule

\multirow{10}{*}{\rotatebox[origin=c]{90}{MLP}}
& BL\_1 & 0.8371 & 0.8111 & 0.8647 & 0.9221 & 0.8433 & 0.8367 & \textbf{0.27} \\
& BL\_2 & 0.8429 & 0.8167 & 0.8706 & 0.9224 & 0.8492 & 0.8423 & 0.28 \\
& BL\_3 & 0.8257 & 0.8000 & 0.8529 & 0.9221 & 0.8330 & 0.8251 & 0.28 \\
& BL\_4 & 0.8314 & 0.8167 & 0.8471 & 0.9178 & 0.8387 & 0.8307 & 0.28 \\
& BL\_5 & 0.8343 & 0.8056 & 0.8647 & 0.9234 & 0.8400 & 0.8339 & 0.29 \\
& BL\_6 & 0.8286 & 0.8000 & 0.8588 & 0.9208 & 0.8344 & 0.8281 & 0.28 \\
& BL\_7 & 0.8314 & 0.8000 & 0.8647 & 0.9194 & 0.8379 & 0.8309 & 0.28 \\
& BL\_8 & 0.8400 & \textbf{0.8222} & 0.8588 & 0.9178 & 0.8445 & 0.8397 & 0.28 \\
& StackMax & 0.8343 & 0.8111 & 0.8588 & 0.9214 & 0.8399 & 0.8338 & N/A \\
& StackMean & 0.8343 & 0.8111 & 0.8588 & 0.9208 & 0.8399 & 0.8338 & N/A \\
\bottomrule
\end{tabular}
\end{table}

\begin{figure}[h!]
\centering
\includegraphics[width=\textwidth]{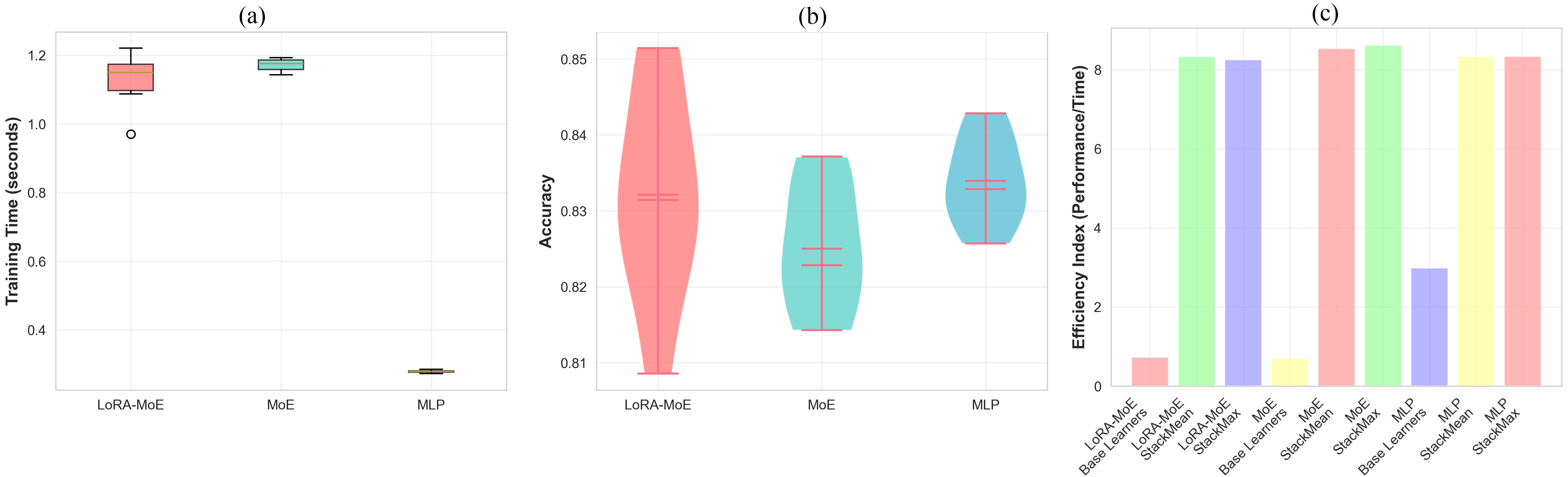}
\caption{Comparison of (a) training time, (b) base-learner accuracy distribution, and (c) efficiency index (performance/time) for LoRA-MoE, MoE, and MLP under different ranks of LoRA.}
\label{fig:lora_rank_efficiency}
\end{figure}

Fig.~\ref{fig:lora_rank_efficiency} provides a visual analysis of training cost, performance stability, and efficiency under different LoRA rank settings. As shown in Fig.~\ref{fig:lora_rank_efficiency}(a), LoRA-MoE maintains lower and more stable training times than the standard MoE, while MLP remains the fastest due to its simpler architecture. The base-learner accuracy distributions in Fig.~\ref{fig:lora_rank_efficiency}(b) indicate that LoRA-MoE achieves consistently higher accuracy with moderate variability, whereas MoE exhibits lower and less stable performance. Finally, the efficiency index in Fig.~\ref{fig:lora_rank_efficiency}(c) clearly favors LoRA-MoE, particularly when combined with stacking strategies, confirming that low-rank adaptation enables an effective balance between accuracy and computational efficiency.

%----------------------------------------------------------
\subsection{Handwriting-Based Independent Task Prediction}
In this experiment, the DARWIN dataset is reformulated by treating each handwriting task as an independent prediction problem, using the 18 extracted features per task. The models are evaluated with hidden dimensions ranging from 5 to 40 (step size 5), corresponding to BL\_1 to BL\_8, while the number of experts for both LoRA-MoE and MoE is fixed to 5. This setting significantly increases the number of samples but removes subject-level aggregation, making the classification problem more challenging and noisier.Table~\ref{tab:model_comparison_hiddendim2} shows that all models experience a clear performance drop compared to the subject-level strategy, indicating that individual handwriting tasks provide weaker and less discriminative information. Nevertheless, both LoRA-MoE and MoE consistently outperform the MLP baseline across most configurations, demonstrating the benefit of expert-based modeling even at the task level. LoRA-MoE achieves competitive performance with fewer hidden dimensions, while MoE benefits more from stacking strategies, achieving the best overall accuracy (69.09\%). These results indicate that expert architectures remain more robust than single-network baselines when task-level variability is high. However, subject-level aggregation remains more effective for reliable Alzheimer’s disease diagnosis.

\begin{table}[h!]
\centering
%\caption{Model Comparison with Different Hidden Dimensions}
\caption{Performance Comparison of Models under Independent Task-Level Prediction.}
\label{tab:model_comparison_hiddendim2}
\small
\begin{tabular}{@{}llccccccc@{}}
\toprule
\textbf{Arch} & \textbf{Model} & \textbf{Accuracy} & \textbf{Sensitivity} & \textbf{Specificity} & \textbf{AUC} & \textbf{Precision} & \textbf{F1 Score} & \textbf{Time(s)} \\
\midrule

\multirow{10}{*}{\rotatebox[origin=c]{90}{LoRA-MoE}}
& BL\_1 & 0.6279 & 0.5854 & 0.6725 & 0.6803 & 0.6315 & 0.6264 & 13.27 \\
& BL\_2 & 0.6577 & 0.6301 & 0.6866 & 0.7112 & 0.6598 & 0.6571 & 13.23 \\
& BL\_3 & 0.6626 & 0.6148 & 0.7127 & 0.7178 & 0.6667 & 0.6615 & 13.47 \\
& BL\_4 & 0.6693 & 0.6189 & 0.7221 & 0.7293 & 0.6741 & 0.6679 & 13.14 \\
& BL\_5 & 0.6784 & 0.6283 & 0.7308 & 0.7321 & 0.6826 & 0.6774 & 14.02 \\
& BL\_6 & 0.6706 & 0.6335 & 0.7094 & 0.7328 & 0.6737 & 0.6698 & 13.91 \\
& BL\_7 & 0.6761 & 0.6308 & 0.7235 & 0.7411 & 0.6824 & 0.6743 & 13.74 \\
& BL\_8 & 0.6793 & 0.6106 & 0.7513 & 0.7448 & 0.6878 & 0.6767 & 14.33 \\
& StackMax & 0.6878 & 0.6369 & 0.7412 & 0.7473 & 0.6916 & 0.6871 & N/A \\
& StackMean & 0.6885 & 0.6360 & 0.7435 & 0.7469 & 0.6926 & 0.6877 & N/A \\
\midrule

\multirow{10}{*}{\rotatebox[origin=c]{90}{MoE}}
& BL\_1 & 0.6383 & 0.5717 & 0.7080 & 0.6927 & 0.6442 & 0.6361 & 10.71 \\
& BL\_2 & 0.6648 & 0.5982 & 0.7346 & 0.7152 & 0.6706 & 0.6632 & 10.83 \\
& BL\_3 & 0.6670 & 0.6088 & 0.7280 & 0.7223 & 0.6714 & 0.6659 & 10.67 \\
& BL\_4 & 0.6770 & 0.6231 & 0.7334 & 0.7302 & 0.6811 & 0.6761 & 10.86 \\
& BL\_5 & 0.6795 & 0.6070 & 0.7555 & 0.7421 & 0.6869 & 0.6775 & 11.81 \\
& BL\_6 & 0.6870 & 0.6220 & 0.7551 & 0.7434 & 0.6938 & 0.6853 & 10.87 \\
& BL\_7 & 0.6890 & 0.6106 & \textbf{0.7711} & 0.7462 & 0.6964 & 0.6872 & 10.85 \\
& BL\_8 & 0.6828 & 0.6092 & 0.7598 & 0.7425 & 0.6895 & 0.6811 & 10.84 \\
& StackMax & 0.6878 & 0.6387 & 0.7393 & 0.7507 & 0.6914 & 0.6871 & N/A \\
& StackMean & \textbf{0.6909} & 0.6222 & 0.7628 & \textbf{0.7533} & \textbf{0.6970} & \textbf{0.6895} & N/A \\
\midrule

\multirow{10}{*}{\rotatebox[origin=c]{90}{MLP}}
& BL\_1 & 0.6211 & 0.6072 & 0.6358 & 0.6658 & 0.6222 & 0.6208 & \textbf{3.53} \\
& BL\_2 & 0.6461 & 0.6043 & 0.6899 & 0.6982 & 0.6496 & 0.6451 & 3.58 \\
& BL\_3 & 0.6498 & 0.6207 & 0.6802 & 0.7003 & 0.6519 & 0.6492 & 3.58 \\
& BL\_4 & 0.6641 & 0.6076 & 0.7233 & 0.7170 & 0.6682 & 0.6630 & 3.58 \\
& BL\_5 & 0.6670 & 0.6198 & 0.7165 & 0.7235 & 0.6720 & 0.6654 & 3.59 \\
& BL\_6 & 0.6690 & \textbf{0.6582} & 0.6802 & 0.7279 & 0.6711 & 0.6682 & 3.72 \\
& BL\_7 & 0.6743 & 0.6211 & 0.7299 & 0.7334 & 0.6804 & 0.6723 & 3.77 \\
& BL\_8 & 0.6777 & 0.6146 & 0.7438 & 0.7364 & 0.6849 & 0.6757 & 3.65 \\
& StackMax & 0.6725 & 0.6299 & 0.7172 & 0.7268 & 0.6758 & 0.6718 & N/A \\
& StackMean & 0.6732 & 0.6335 & 0.7148 & 0.7278 & 0.6762 & 0.6726 & N/A \\
\bottomrule
\end{tabular}
\end{table}

\subsection{Impact of Model Depth (Multi-Layer Architecture)}
This experiment analyzes the effect of increasing model depth by evaluating deeper architectures for both MLP and expert networks. Five-layer and eight-layer configurations are considered, with Table~\ref{tab:model_comparison_hiddendim_5layer} reporting the results for the five-layer setting. The hidden dimension varies from 50 to 400 (BL\_1 to BL\_8, step size 50), and the number of experts for LoRA-MoE and MoE is fixed to 6. As shown in Table~\ref{tab:model_comparison_hiddendim_5layer}, increasing network depth does not lead to consistent performance improvements. LoRA-MoE remains competitive but does not surpass its shallower version, while the standard MoE suffers from performance degradation and increased training time due to higher optimization complexity. In contrast, the MLP achieves similar or better accuracy with significantly lower computational cost. These results suggest that moderate-depth architectures are sufficient for handwriting-based Alzheimer’s disease diagnosis, and deeper models mainly introduce unnecessary complexity.

\begin{table}[h!]
\centering
\caption{Model Comparison with Different Hidden Dimensions(5 Layers)}
\label{tab:model_comparison_hiddendim_5layer}
\small
\begin{tabular}{@{}llccccccc@{}}
\toprule
\textbf{Arch} & \textbf{Model} & \textbf{Accuracy} & \textbf{Sensitivity} & \textbf{Specificity} & \textbf{AUC} & \textbf{Precision} & \textbf{F1 Score} & \textbf{Time(s)} \\
\midrule

\multirow{10}{*}{\rotatebox[origin=c]{90}{LoRA-MoE}}
& BL\_1 & 0.8314 & 0.8278 & 0.8353 & 0.9162 & 0.8358 & 0.8310 & 1.72 \\
& BL\_2 & 0.8143 & 0.8167 & 0.8118 & 0.9041 & 0.8203 & 0.8136 & 1.77 \\
& BL\_3 & 0.8457 & 0.8500 & 0.8412 & 0.8938 & 0.8506 & 0.8452 & 1.98 \\
& BL\_4 & 0.8371 & 0.8333 & 0.8412 & 0.8946 & 0.8405 & 0.8368 & 2.12 \\
& BL\_5 & 0.8286 & 0.8389 & 0.8176 & 0.8923 & 0.8305 & 0.8283 & 2.29 \\
& BL\_6 & 0.8371 & \textbf{0.8722} & 0.8000 & 0.8869 & 0.8411 & 0.8365 & 2.92 \\
& BL\_7 & 0.8314 & 0.8556 & 0.8059 & 0.9124 & 0.8353 & 0.8304 & 3.15 \\
& BL\_8 & 0.8143 & 0.8389 & 0.7882 & 0.8812 & 0.8174 & 0.8137 & 3.23 \\
& StackMax & 0.8400 & 0.8333 & 0.8471 & 0.9183 & 0.8437 & 0.8396 & N/A \\
& StackMean & 0.8429 & 0.8500 & 0.8353 & 0.9181 & 0.8453 & 0.8426 & N/A \\
\midrule

\multirow{10}{*}{\rotatebox[origin=c]{90}{MoE}}
& BL\_1 & 0.8229 & 0.8000 & 0.8471 & 0.9059 & 0.8290 & 0.8221 & 1.56 \\
& BL\_2 & 0.7971 & 0.7611 & 0.8353 & 0.9011 & 0.8046 & 0.7959 & 1.96 \\
& BL\_3 & 0.8143 & 0.8111 & 0.8176 & 0.8948 & 0.8159 & 0.8141 & 2.64 \\
& BL\_4 & 0.7743 & 0.7889 & 0.7588 & 0.8750 & 0.7816 & 0.7727 & 2.71 \\
& BL\_5 & 0.8143 & 0.7667 & 0.8647 & 0.8889 & 0.8230 & 0.8134 & 3.39 \\
& BL\_6 & 0.7714 & 0.7667 & 0.7765 & 0.8513 & 0.7773 & 0.7700 & 4.17 \\
& BL\_7 & 0.8029 & 0.8000 & 0.8059 & 0.8721 & 0.8080 & 0.8022 & 4.57 \\
& BL\_8 & 0.8086 & 0.7889 & 0.8294 & 0.8675 & 0.8138 & 0.8080 & 4.82 \\
& StackMax & 0.8171 & 0.6944 & \textbf{0.9471} & 0.9201 & \textbf{0.8457} & 0.8138 & N/A \\
& StackMean & 0.8371 & 0.8111 & 0.8647 & \textbf{0.9252} & 0.8437 & 0.8367 & N/A \\
\midrule

\multirow{10}{*}{\rotatebox[origin=c]{90}{MLP}}
& BL\_1 & 0.8486 & 0.8556 & 0.8412 & 0.9212 & 0.8506 & 0.8484 & \textbf{0.39} \\
& BL\_2 & 0.8486 & 0.8722 & 0.8235 & 0.8993 & 0.8524 & 0.8479 & 0.41 \\
& BL\_3 & \textbf{0.8543} & 0.8667 & 0.8412 & 0.9020 & \textbf{0.8556} & \textbf{0.8541} & 0.53 \\
& BL\_4 & 0.8429 & 0.8556 & 0.8294 & 0.9023 & 0.8473 & 0.8424 & 0.57 \\
& BL\_5 & 0.8343 & 0.8389 & 0.8294 & 0.8922 & 0.8389 & 0.8338 & 0.69 \\
& BL\_6 & 0.8371 & 0.8444 & 0.8294 & 0.8899 & 0.8439 & 0.8363 & 0.87 \\
& BL\_7 & 0.8314 & 0.8333 & 0.8294 & 0.8887 & 0.8405 & 0.8299 & 0.99 \\
& BL\_8 & 0.8143 & 0.8444 & 0.7824 & 0.8788 & 0.8211 & 0.8131 & 0.99 \\
& StackMax & 0.8457 & 0.8500 & 0.8412 & 0.9191 & 0.8490 & 0.8454 & N/A \\
& StackMean & 0.8514 & 0.8500 & 0.8529 & 0.9155 & 0.8546 & 0.8512 & N/A \\

\bottomrule
\end{tabular}
\end{table}

\medskip
A similar trend is observed in the eight-layer setting (Table~\ref{tab:model_comparison_hiddendim_8layer}). Although stacking strategies improve LoRA-MoE performance, the overall gains remain marginal compared to shallower architectures, while training time increases sharply for both LoRA-MoE and MoE. In contrast, MLP maintains stable performance with moderate computational growth. Overall, these results indicate that increasing model depth provides limited benefit for handwriting-based Alzheimer’s disease diagnosis. Moderate-depth architectures are sufficient to capture relevant cognitive–motor patterns, and deeper models mainly introduce unnecessary complexity and computational overhead without reliable performance gains.

\begin{table}[h!]
\centering
\caption{Model Comparison with Different Hidden Dimensions(8 Layers)}
\label{tab:model_comparison_hiddendim_8layer}
\small
\begin{tabular}{@{}llccccccc@{}}
\toprule
\textbf{Arch} & \textbf{Model} & \textbf{Accuracy} & \textbf{Sensitivity} & \textbf{Specificity} & \textbf{AUC} & \textbf{Precision} & \textbf{F1 Score} & \textbf{Time(s)} \\
\midrule

\multirow{10}{*}{\rotatebox[origin=c]{90}{LoRA-MoE}}
& BL\_1 & 0.8386 & 0.8556 & 0.8206 & 0.8955 & 0.8458 & 0.8374 & 4.04 \\
& BL\_2 & 0.8429 & 0.8444 & 0.8412 & 0.8856 & 0.8518 & 0.8410 & 4.33 \\
& BL\_3 & 0.8543 & 0.8861 & 0.8206 & 0.8907 & 0.8630 & 0.8531 & 4.90 \\
& BL\_4 & 0.8429 & 0.8667 & 0.8176 & 0.8796 & 0.8514 & 0.8417 & 5.24 \\
& BL\_5 & 0.8400 & 0.8611 & 0.8176 & 0.8834 & 0.8457 & 0.8392 & 5.79 \\
& BL\_6 & 0.8329 & 0.8833 & 0.7794 & 0.8852 & 0.8430 & 0.8312 & 7.60 \\
& BL\_7 & 0.8229 & 0.8583 & 0.7853 & 0.8718 & 0.8318 & 0.8208 & 8.12 \\
& BL\_8 & 0.8286 & 0.8500 & 0.8059 & 0.8854 & 0.8381 & 0.8269 & 8.42 \\
& StackMax & 0.8386 & 0.8028 & 0.8765 & 0.9165 & 0.8490 & 0.8371 & N/A \\
& StackMean & \textbf{0.8600} & \textbf{0.8889} & 0.8294 & \textbf{0.9193} & \textbf{0.8670} & \textbf{0.8590} & N/A \\
\midrule

\multirow{10}{*}{\rotatebox[origin=c]{90}{MoE}}
& BL\_1 & 0.8071 & 0.8056 & 0.8088 & 0.8935 & 0.8149 & 0.8057 & \textbf{3.58} \\
& BL\_2 & 0.8257 & 0.8472 & 0.8029 & 0.8771 & 0.8343 & 0.8244 & 4.80 \\
& BL\_3 & 0.7957 & 0.8111 & 0.7794 & 0.8303 & 0.8019 & 0.7944 & 6.66 \\
& BL\_4 & 0.7886 & 0.8000 & 0.7765 & 0.8326 & 0.7952 & 0.7869 & 7.00 \\
& BL\_5 & 0.7886 & 0.7917 & 0.7853 & 0.8127 & 0.7957 & 0.7872 & 8.64 \\
& BL\_6 & 0.7986 & 0.8167 & 0.7794 & 0.8261 & 0.8048 & 0.7974 & 10.88 \\
& BL\_7 & 0.7871 & 0.7556 & 0.8206 & 0.8215 & 0.7954 & 0.7855 & 11.79 \\
& BL\_8 & 0.7614 & 0.7639 & 0.7588 & 0.7922 & 0.7702 & 0.7592 & 12.51 \\
& StackMax & 0.7414 & 0.5444 & \textbf{0.9500} & 0.8624 & 0.8011 & 0.7289 & N/A \\
& StackMean & 0.8471 & 0.8583 & 0.8353 & 0.9150 & 0.8517 & 0.8465 & N/A \\
\midrule

\multirow{10}{*}{\rotatebox[origin=c]{90}{MLP}}
& BL\_1 & 0.8486 & 0.8556 & 0.8412 & 0.9047 & 0.8554 & 0.8479 & 0.80 \\
& BL\_2 & 0.8429 & 0.8583 & 0.8265 & 0.8976 & 0.8492 & 0.8420 & 0.96 \\
& BL\_3 & 0.8386 & 0.8528 & 0.8235 & 0.8931 & 0.8461 & 0.8374 & 1.28 \\
& BL\_4 & 0.8400 & 0.8833 & 0.7941 & 0.8857 & 0.8471 & 0.8386 & 1.37 \\
& BL\_5 & 0.8357 & 0.8500 & 0.8206 & 0.8879 & 0.8422 & 0.8346 & 1.71 \\
& BL\_6 & 0.8386 & 0.8806 & 0.7941 & 0.8855 & 0.8492 & 0.8366 & 2.18 \\
& BL\_7 & 0.8357 & 0.8750 & 0.7941 & 0.8880 & 0.8455 & 0.8337 & 2.46 \\
& BL\_8 & 0.8343 & 0.8583 & 0.8088 & 0.8948 & 0.8442 & 0.8328 & 2.62 \\
& StackMax & 0.8571 & 0.8528 & 0.8618 & 0.9106 & 0.8647 & 0.8562 & N/A \\
& StackMean & 0.8500 & 0.8694 & 0.8294 & 0.9114 & 0.8552 & 0.8492 & N/A \\

\bottomrule
\end{tabular}
\end{table}

%-----------------------------------------------------------
\subsection{Classification by Task} 
 In this experiment, the DARWIN dataset is decomposed at the task level into 25 subsets, each corresponding to the 18 features associated with a specific task, as detailed in Table~\ref{tab:handwriting_features}. For each model configuration, a separate model is trained on each subset, yielding independent classification results for each task. The predictions from the 25 models are then aggregated via a hard voting strategy, where the class with the highest frequency across tasks is selected as the final output for each subject.

For each subset, we trained 5 instances of 3 distinct architectures—MLP, MoE, and LoRA-MoE, with hidden layer sizes varying from 5 to 25 by step 5. For each architecture, the 5 trained models are further combined using both mean and max stacking strategies. All models shared a fixed depth of 3 layers. For the MoE and LoRA-MoE variants, the number of experts is set to 5, with a top‑k value of 1 (i.e., only the highest-scoring expert is activated per sample). In the LoRA-MoE configuration, both the rank of each LoRA adapter and the scaling factor $\alpha$ are fixed to 1.

To ensure the reliability of the results, each experimental setting is repeated 20 times. In each repetition, the complete DARWIN dataset is first randomly split into training and test sets, and then the task-level decomposition is applied. This order ensures that the same subject's data across different tasks is consistently assigned to either the training or test fold, thereby preserving the validity of the voting ensemble evaluation. Table~\ref{tab:lr_accuracy} shows the mean accuracy of each model of LoRA-MoE architecture on each task, and Table~\ref{tab:voting_ensembles} shows the results of multi-task voting ensemble. More details about the performance of each model for individual handwriting tasks can be found in Appendix A.

For the performance of each model in AD diagnosis based on individual tasks, Table~\ref{tab:lr_accuracy}, Table~\ref{tab:mlp_accuracy}, Table~\ref{tab:moe_accuracy} present the accuracy results. All 3 architectures achieve their optimal performance in Exp 17, with LoRA-MoE attaining the highest accuracy (MLP-Mean: 79.29\%, MoE-Max: 79.00\%, LM-Max: 80.29\%). Table~\ref{tab:lm_sensitivity}, Table~\ref{tab:mlp_sensitivity}, Table~\ref{tab:moe_sensitivity} report sensitivity, where LoRA-MoE also demonstrates superior performance (Exp 17 LM-Max: 79.72\%, Exp 7 MLP-5: 79.17\%, Exp 17 MoE-Max: 80.00\%). Table~\ref{tab:lm_specificity}, Table~\ref{tab:mlp_specificity}, Table~\ref{tab:moe_specificity} display specificity; while LoRA-MoE lags behind MLP, it still outperforms MoE (Exp 9 LM-Mean: 84.41\%, Exp 8 MLP-15: 86.18\%, Exp 8 MoE-Max: 80.88\%). Collectively, these results indicate that even under the challenging condition of single-task prediction with substantially reduced information, LoRA-MoE still yields competitive outcomes.
\begin{table}[h!]
\centering
\caption{Accuracy (\%) of LM(LoRA-MoE) configurations on each experiment.}
\label{tab:lr_accuracy}
\small
\begin{tabular}{lccccccc}
\toprule
\textbf{Exp} & \textbf{LM-5} & \textbf{LM-10} & \textbf{LM-15} & \textbf{LM-20} & \textbf{LM-25} & \textbf{LM-Max} & \textbf{LM-Mean} \\
\midrule
1 & 60.43 & 62.14 & 64.00 & 62.29 & 61.14 & 61.71 & 63.71 \\
2 & 63.14 & 68.14 & 66.71 & 69.00 & 68.86 & 68.29 & 69.86 \\
3 & 61.29 & 67.29 & 67.86 & 68.86 & 69.29 & 70.00 & 69.57 \\
4 & 63.71 & 66.29 & 65.57 & 66.29 & 65.86 & 66.00 & 66.57 \\
5 & 66.57 & 69.43 & 72.00 & 72.71 & 70.00 & 71.57 & 72.00 \\
6 & 67.57 & 69.29 & 69.14 & 69.71 & 67.86 & 70.86 & 70.57 \\
7 & 74.14 & 74.14 & 73.14 & 74.14 & 74.43 & 75.57 & 75.57 \\
8 & 70.43 & 71.86 & 73.00 & 74.29 & 73.14 & 73.43 & 74.14 \\
9 & 71.00 & 74.43 & 75.57 & 73.71 & 74.29 & 74.57 & 75.00 \\
10 & 68.14 & 68.86 & 68.86 & 70.57 & 70.29 & 71.00 & 70.86 \\
11 & 64.43 & 66.00 & 66.86 & 68.00 & 68.57 & 68.86 & 67.71 \\
12 & 62.86 & 67.43 & 68.00 & 68.43 & 71.29 & 70.43 & 70.14 \\
13 & 64.43 & 67.14 & 66.57 & 66.86 & 66.86 & 67.14 & 67.71 \\
14 & 61.71 & 64.29 & 68.29 & 69.71 & 69.71 & 70.29 & 70.29 \\
15 & 65.14 & 72.71 & 71.71 & 73.00 & 73.14 & 73.29 & 73.71 \\
16 & 66.29 & 74.00 & 74.57 & 74.43 & 75.29 & 75.43 & 75.57 \\
17 & 73.43 & 77.57 & 77.71 & 76.29 & 77.00 & 80.29 & 79.86 \\
18 & 62.57 & 68.29 & 68.43 & 70.43 & 70.71 & 70.29 & 69.86 \\
19 & 68.00 & 71.86 & 75.14 & 72.29 & 74.43 & 74.57 & 75.86 \\
20 & 66.14 & 68.71 & 69.29 & 70.14 & 70.14 & 70.86 & 70.29 \\
21 & 65.86 & 69.57 & 68.86 & 69.57 & 72.71 & 69.43 & 69.29 \\
22 & 67.00 & 70.86 & 70.29 & 70.14 & 70.86 & 70.43 & 70.29 \\
23 & 71.43 & 75.29 & 76.00 & 73.71 & 74.29 & 74.86 & 75.14 \\
24 & 63.71 & 70.86 & 74.14 & 72.00 & 72.86 & 74.71 & 75.43 \\
25 & 67.86 & 71.00 & 71.14 & 71.14 & 69.57 & 72.43 & 73.00 \\
\bottomrule
\end{tabular}
\end{table}

Table~\ref{tab:voting_ensembles} summarizes the performance of each configuration following voting ensemble. Compared to independent single-task prediction, the voting strategy enhances the performance of most models. The ensemble model based on MLP-25 achieves the best overall results in terms of accuracy (87.43\%), sensitivity (84.17\%), and specificity (90.88\%), demonstrating that task-level independent training followed by a voting strategy can effectively improve overall diagnostic performance. The LM-25 ensemble model for LoRA-MoE (accuracy: 86.86\%, sensitivity: 83.33\%, specificity: 90.59\%) ranks second, indicating that LoRA-MoE maintains strong capability while preserving high specificity. The MoE-25 ensemble model (accuracy: 85.00\%, sensitivity: 81.67\%, specificity: 88.53\%) exhibits lower performance than the former two. However, under the 2 stacking configurations, MoE-Mean and MoE-Max outperform LoRA-MoE, particularly in accuracy and sensitivity, while MLP exhibits the weakest performance in these settings. This suggests that the single fully-connected architecture of MLP, when faced with diverse tasks, cannot capture inter-task variations simply by increasing hidden dimensions, due to its lack of specialization.

Notably, under the configuration with the smallest hidden dimension (i.e., =5), LoRA-MoE (LM-5) surpasses both MoE and MLP across accuracy, sensitivity, and specificity. Given that top‑k is set to 1 and the LoRA adapter rank is 1, the number of activated parameters during the inference phase of LoRA-MoE is substantially smaller than that of MoE and MLP. This finding underscores that LoRA-MoE achieves superior performance compared to standard MoE and MLP while utilizing significantly fewer activated parameters, thereby highlighting its strong AD identification capacity in computationally-constrained scenarios.

\begin{table}[h!]
\centering
\caption{Performance of Voting Ensemble configurations (mean ± std, \%).}
\label{tab:voting_ensembles}
\small
\begin{tabular}{lcccccc}
\toprule
\textbf{Configuration} & \textbf{Accuracy (\%)} & \textbf{Sensitivity (\%)} & \textbf{Specificity (\%)} \\
\midrule
\textbf{MLP} \\
MLP-5 & 80.14 ± 6.61 & 81.67 ± 14.07 & 78.53 ± 9.71 \\
MLP-10 & 81.71 ± 6.48 & 79.17 ± 10.23 & 84.41 ± 7.03 \\
MLP-15 & 82.00 ± 7.18 & 76.11 ± 11.12 & 88.24 ± 5.26 \\
MLP-20 & 84.71 ± 5.21 & 80.28 ± 9.21 & 89.41 ± 5.13 \\
MLP-25 & 87.43 ± 4.64 & 84.17 ± 7.51 & 90.88 ± 5.09 \\
MLP-Mean & 84.57 ± 5.45 & 80.00 ± 9.69 & 89.41 ± 4.78 \\
MLP-Max & 84.43 ± 4.99 & 80.83 ± 8.33 & 88.24 ± 5.26 \\
\midrule
\textbf{MoE} \\
MoE-5 & 79.86 ± 6.85 & 80.00 ± 11.97 & 79.71 ± 9.75 \\
MoE-10 & 84.86 ± 5.72 & 83.89 ± 8.59 & 85.88 ± 6.28 \\
MoE-15 & 86.00 ± 3.93 & 86.94 ± 6.40 & 85.00 ± 6.30 \\
MoE-20 & 85.43 ± 5.02 & 84.72 ± 7.22 & 86.18 ± 5.35 \\
MoE-25 & 85.00 ± 6.19 & 81.67 ± 10.26 & 88.53 ± 6.02 \\
MoE-Mean & 86.14 ± 4.63 & 85.83 ± 9.04 & 86.47 ± 3.77 \\
MoE-Max & 86.14 ± 3.86 & 85.56 ± 7.54 & 86.76 ± 5.22 \\
\midrule
\textbf{LoRA-MoE} \\
LM-5 & 83.00 ± 7.31 & 83.89 ± 12.29 & 82.06 ± 8.82 \\
LM-10 & 82.71 ± 5.53 & 78.61 ± 9.67 & 87.06 ± 4.40 \\
LM-15 & 84.29 ± 5.53 & 80.83 ± 8.33 & 87.94 ± 5.42 \\
LM-20 & 85.43 ± 5.99 & 82.50 ± 8.29 & 88.53 ± 5.42 \\
LM-25 & 86.86 ± 5.74 & 83.33 ± 7.66 & 90.59 ± 5.70 \\
LM-Mean & 85.57 ± 5.15 & 82.22 ± 8.89 & 89.12 ± 5.96 \\
LM-Max & 85.86 ± 4.82 & 83.89 ± 7.01 & 87.94 ± 5.73 \\
\bottomrule
\end{tabular}
\end{table}

\newpage
\medskip
Overall, the experimental results highlight several advantages of the proposed LoRA-MoE framework for handwriting-based Alzheimer’s disease diagnosis. By combining expert specialization with low-rank parameter adaptation, the model achieves competitive or superior performance while activating significantly fewer parameters during inference. This property makes LoRA-MoE particularly attractive for computationally constrained clinical screening systems and portable diagnostic platforms. However, several challenges remain. First, handwriting-based diagnosis is inherently sensitive to inter-subject variability and task-specific noise, which can affect single-task predictions. Second, the relatively limited size of clinical datasets such as DARWIN may restrict the ability of complex expert architectures to fully generalize. Potential solutions include incorporating larger multi-center handwriting datasets, integrating multimodal cognitive biomarkers, and developing adaptive expert routing strategies that dynamically select informative tasks. Such directions could further improve the robustness, interpretability, and clinical applicability of handwriting-based Alzheimer’s disease diagnosis systems.

%\newpage
\section{Conclusion}\label{sec4}
This study introduced a novel Low-Rank Mixture of Experts (LoRA-MoE) architecture for Alzheimer’s disease diagnosis based on handwriting data. By integrating low-rank adaptation into a shared expert framework, the proposed model achieves effective expert specialization while significantly reducing model complexity and training cost. Extensive experiments on the DARWIN dataset demonstrate that LoRA-MoE consistently outperforms standard MoE and MLP baselines across multiple evaluation settings. A comprehensive ablation analysis showed that LoRA-MoE benefits most from moderate hidden dimensions, a limited number of experts, and low-rank configurations, confirming that carefully constrained model capacity is crucial for reliable performance on small and high-dimensional clinical datasets. Increasing either the number of experts, the LoRA rank, or the network depth beyond moderate values leads to diminishing returns and higher computational overhead. Moreover, stacking strategies further improve robustness and efficiency by aggregating complementary expert predictions. Additional experiments under independent task-level classification highlighted the robustness of expert-based architectures in challenging, noisy settings, although subject-level aggregation remains more effective for reliable diagnosis. Overall, the proposed LoRA-MoE framework offers a favorable balance between accuracy, robustness, and computational efficiency, making it well-suited for practical handwriting-based Alzheimer’s disease screening systems.

Future work will extend LoRA-MoE to include multimodal cognitive biomarkers, such as speech and drawing dynamics, and investigate adaptive routing strategies and automated rank selection to further enhance scalability and clinical applicability.

\bibliographystyle{unsrt}  
\bibliography{references}

@article{kavitha2022early,
  title={Early-stage Alzheimer's disease prediction using machine learning models},
  author={Kavitha, C and Mani, Vinodhini and Srividhya, SR and Khalaf, Osamah Ibrahim and Tavera Romero, Carlos Andres},
  journal={Frontiers in public health},
  volume={10},
  pages={853294},
  year={2022},
  publisher={Frontiers Media SA}
}

@article{kruthika2019multistage,
  title={Multistage classifier-based approach for Alzheimer's disease prediction and retrieval},
  author={Kruthika, KR and Maheshappa, HD and Alzheimer's Disease Neuroimaging Initiative and others},
  journal={Informatics in Medicine Unlocked},
  volume={14},
  pages={34--42},
  year={2019},
  publisher={Elsevier}
}

@article{jack2018nia,
  title={NIA-AA research framework: toward a biological definition of Alzheimer's disease},
  author={Jack Jr, Clifford R and Bennett, David A and Blennow, Kaj and Carrillo, Maria C and Dunn, Billy and Haeberlein, Samantha Budd and Holtzman, David M and Jagust, William and Jessen, Frank and Karlawish, Jason and others},
  journal={Alzheimer's \& dementia},
  volume={14},
  number={4},
  pages={535--562},
  year={2018},
  publisher={Wiley Online Library}
}

@article{scheltens2021alzheimer,
  title={Alzheimer's disease},
  author={Scheltens, Philip and De Strooper, Bart and Kivipelto, Miia and Holstege, Henne and Ch{\'e}telat, Gael and Teunissen, Charlotte E and Cummings, Jeffrey and van der Flier, Wiesje M},
  journal={The Lancet},
  volume={397},
  number={10284},
  pages={1577--1590},
  year={2021},
  publisher={Elsevier}
}

@article{blennow2015clinical,
  title={Clinical utility of cerebrospinal fluid biomarkers in the diagnosis of early Alzheimer's disease},
  author={Blennow, Kaj and Dubois, Bruno and Fagan, Anne M and Lewczuk, Piotr and De Leon, Mony J and Hampel, Harald},
  journal={Alzheimer's \& Dementia},
  volume={11},
  number={1},
  pages={58--69},
  year={2015},
  publisher={Elsevier}
}

@article{weiner2013alzheimer,
  title={The Alzheimer's Disease Neuroimaging Initiative: a review of papers published since its inception},
  author={Weiner, Michael W and Veitch, Dallas P and Aisen, Paul S and Beckett, Laurel A and Cairns, Nigel J and Green, Robert C and Harvey, Danielle and Jack, Clifford R and Jagust, William and Liu, Enchi and others},
  journal={Alzheimer's \& Dementia},
  volume={9},
  number={5},
  pages={e111--e194},
  year={2013},
  publisher={Elsevier}
}

@article{wang2020early,
  title={Early detection of Parkinson’s disease using deep learning and machine learning},
  author={Wang, Wu and Lee, Junho and Harrou, Fouzi and Sun, Ying},
  journal={IEEE access},
  volume={8},
  pages={147635--147646},
  year={2020},
  publisher={IEEE}
}

@article{harrou2025manifold,
  title={A Manifold Learning-Based Anomaly Detection Framework for Cardiovascular Disease Diagnosis},
  author={Harrou, Fouzi and Dairi, Abdelkader and Sun, Ying},
  journal={Computational Intelligence},
  volume={41},
  number={5},
  pages={e70130},
  year={2025},
  publisher={Wiley Online Library}
}

@article{khandakar2021machine,
  title={A machine learning model for early detection of diabetic foot using thermogram images},
  author={Khandakar, Amith and Chowdhury, Muhammad EH and Reaz, Mamun Bin Ibne and Ali, Sawal Hamid Md and Hasan, Md Anwarul and Kiranyaz, Serkan and Rahman, Tawsifur and Alfkey, Rashad and Bakar, Ahmad Ashrif A and Malik, Rayaz A},
  journal={Computers in biology and medicine},
  volume={137},
  pages={104838},
  year={2021},
  publisher={Elsevier}
}

@article{litjens2017survey,
  title={A survey on deep learning in medical image analysis},
  author={Litjens, Geert and Kooi, Thijs and Bejnordi, Babak Ehteshami and Setio, Arnaud Arindra Adiyoso and Ciompi, Francesco and Ghafoorian, Mohsen and Van Der Laak, Jeroen Awm and Van Ginneken, Bram and S{\'a}nchez, Clara I},
  journal={Medical image analysis},
  volume={42},
  pages={60--88},
  year={2017},
  publisher={Elsevier}
}

@article{esteva2019guide,
  title={A guide to deep learning in healthcare},
  author={Esteva, Andre and Robicquet, Alexandre and Ramsundar, Bharath and Kuleshov, Volodymyr and DePristo, Mark and Chou, Katherine and Cui, Claire and Corrado, Greg and Thrun, Sebastian and Dean, Jeff},
  journal={Nature medicine},
  volume={25},
  number={1},
  pages={24--29},
  year={2019},
  publisher={Nature Publishing Group US New York}
}

@article{wang2024stacked,
  title={Stacked deep learning approach for efficient SARS-CoV-2 detection in blood samples},
  author={Wang, Wu and Harrou, Fouzi and Dairi, Abdelkader and Sun, Ying},
  journal={Artificial Intelligence in Medicine},
  volume={148},
  pages={102767},
  year={2024},
  publisher={Elsevier}
}

@article{fraser2015linguistic,
  title={Linguistic features identify Alzheimer’s disease in narrative speech},
  author={Fraser, Kathleen C and Meltzer, Jed A and Rudzicz, Frank},
  journal={Journal of Alzheimer’s disease},
  volume={49},
  number={2},
  pages={407--422},
  year={2015},
  publisher={SAGE Publications Sage UK: London, England}
}

@article{neils2006dysgraphia,
  title={Dysgraphia in Alzheimer’s disease: a review for clinical and research purposes},
  author={Neils-Strunjas, Jean and Groves-Wright, Kathy and Mashima, Pauline and Harnish, Stacy},
  journal={Journal of speech, Language, and Hearing research},
  volume={49},
  number={6},
  pages={1313--1330},
  year={2006}
}

@article{de2019handwriting,
  title={Handwriting analysis to support neurodegenerative diseases diagnosis: A review},
  author={De Stefano, Claudio and Fontanella, Francesco and Impedovo, Donato and Pirlo, Giuseppe and Di Freca, Alessandra Scotto},
  journal={Pattern Recognition Letters},
  volume={121},
  pages={37--45},
  year={2019},
  publisher={Elsevier}
}

@misc{luz2021alzheimer,
  title={Alzheimer's dementia recognition through spontaneous speech},
  author={Luz, Saturnino and Haider, Fasih and de la Fuente Garcia, Sofia and Fromm, Davida and MacWhinney, Brian},
  journal={Frontiers in computer science},
  volume={3},
  pages={780169},
  year={2021},
  publisher={Frontiers Media SA}
}

@ARTICLE{jra1991moe,
  author={Jacobs, Robert A. and Jordan, Michael I. and Nowlan, Steven J. and Hinton, Geoffrey E.},
  journal={Neural Computation}, 
  title={Adaptive Mixtures of Local Experts}, 
  year={1991},
  volume={3},
  number={1},
  pages={79-87},
 }

@INPROCEEDINGS{jra1993moe,
  author={Jordan, M.I. and Jacobs, R.A.},
  booktitle={Proceedings of 1993 International Conference on Neural Networks (IJCNN-93-Nagoya, Japan)}, 
  title={Hierarchical mixtures of experts and the EM algorithm}, 
  year={1993},
  volume={2},
  number={},
  pages={1339-1344 vol.2},
}

@article{Shazeer2017OutrageouslyLN,
  title={Outrageously Large Neural Networks: The Sparsely-Gated Mixture-of-Experts Layer},
  author={Noam Shazeer and Azalia Mirhoseini and Krzysztof Maziarz and Andy Davis and Quoc V. Le and Geoffrey E. Hinton and Jeff Dean},
  year={2017},
  journal={arXiv preprint arXiv:1701.06538},
}

@article{Lepikhin2020GShardSG,
  title={GShard: Scaling Giant Models with Conditional Computation and Automatic Sharding},
  author={Dmitry Lepikhin and HyoukJoong Lee and Yuanzhong Xu and Dehao Chen and Orhan Firat and Yanping Huang and Maxim Krikun and Noam Shazeer and Z. Chen},
  year={2020},
  journal={arXiv preprint arXiv:2006.16668},
}

@article{fedus2022switch,
  author  = {William Fedus and Barret Zoph and Noam Shazeer},
  title   = {Switch Transformers: Scaling to Trillion Parameter Models with Simple and Efficient Sparsity},
  journal = {Journal of Machine Learning Research},
  year    = {2022},
  volume  = {23},
  number  = {120},
  pages   = {1--39},
}

@inproceedings{Dai2024DeepSeekMoETU,
  title={DeepSeekMoE: Towards Ultimate Expert Specialization in Mixture-of-Experts Language Models},
  author={Damai Dai and Chengqi Deng and Chenggang Zhao and Runxin Xu and Huazuo Gao and Deli Chen and Jiashi Li and Wangding Zeng and Xingkai Yu and Yu Wu and Zhenda Xie and Y. K. Li and Panpan Huang and Fuli Luo and Chong Ruan and Zhifang Sui and Wenfeng Liang},
  booktitle={Annual Meeting of the Association for Computational Linguistics},
  year={2024}
}

@article{Jiang2024MixtralOE,
  title={Mixtral of Experts},
  author={Albert Q. Jiang and Alexandre Sablayrolles and Antoine Roux and Arthur Mensch and Blanche Savary and Chris Bamford and Devendra Singh Chaplot and Diego de Las Casas and Emma Bou Hanna and Florian Bressand and Gianna Lengyel and Guillaume Bour and Guillaume Lample and L{\'e}lio Renard Lavaud and Lucile Saulnier and Marie-Anne Lachaux and Pierre Stock and Sandeep Subramanian and Sophia Yang and Szymon Antoniak and Teven Le Scao and Th{\'e}ophile Gervet and Thibaut Lavril and Thomas Wang and Timoth{\'e}e Lacroix and William El Sayed},
  year={2024},
  journal={arXiv preprint arXiv:2401.04088},
}

@inproceedings{hu2022lora,
author = {Hu, Edward J. and Shen, Yelong and Wallis, Phillip and Allen-Zhu, Zeyuan and Yuanzhi  Li and Wang, Shean and Wang, Lu and Chen, Weizhu},
title = {LoRA: Low-Rank Adaptation of Large Language Models},
booktitle = {ICLR 2022},
year = {2022},
month = {April},
}

@ARTICLE{lialin2023relora,
       author = {{Lialin}, Vladislav and {Shivagunde}, Namrata and {Muckatira}, Sherin and {Rumshisky}, Anna},
        title = "{ReLoRA: High-Rank Training Through Low-Rank Updates}",
      journal = {arXiv preprint arXiv:2307.05695},
         year = 2023,
}

@article{zhang2023adalora,
      title={AdaLoRA: Adaptive Budget Allocation for Parameter-Efficient Fine-Tuning}, 
      author={Qingru Zhang and Minshuo Chen and Alexander Bukharin and Nikos Karampatziakis and Pengcheng He and Yu Cheng and Weizhu Chen and Tuo Zhao},
      year={2023},
      journal={arXiv preprint arXiv:2303.10512},
      archivePrefix={arXiv},
      primaryClass={cs.CL},
}

@inproceedings{dt2023qlora,
 author = {Dettmers, Tim and Pagnoni, Artidoro and Holtzman, Ari and Zettlemoyer, Luke},
 booktitle = {Advances in Neural Information Processing Systems},
 pages = {10088--10115},
 publisher = {Curran Associates, Inc.},
 title = {QLoRA: Efficient Finetuning of Quantized LLMs},
 volume = {36},
 year = {2023}
}

@InProceedings{pmlr-v202-malladi23a,
  title = 	 {A Kernel-Based View of Language Model Fine-Tuning},
  author =       {Malladi, Sadhika and Wettig, Alexander and Yu, Dingli and Chen, Danqi and Arora, Sanjeev},
  booktitle = 	 {Proceedings of the 40th International Conference on Machine Learning},
  pages = 	 {23610--23641},
  year = 	 {2023},
  volume = 	 {202},
  series = 	 {Proceedings of Machine Learning Research},
  month = 	 {23--29 Jul},
  publisher =    {PMLR},
}

@article{zeng2024expressive,
      title={The Expressive Power of Low-Rank Adaptation}, 
      author={Yuchen Zeng and Kangwook Lee},
      year={2024},
      journal={arXiv preprint arXiv:2310.17513},
      archivePrefix={arXiv},
      primaryClass={cs.LG},
}

@article{Huang2023LoraHubEC,
  title={LoraHub: Efficient Cross-Task Generalization via Dynamic LoRA Composition},
  author={Chengsong Huang and Qian Liu and Bill Yuchen Lin and Tianyu Pang and Chao Du and Min Lin},
  year={2023},
  journal={arXiv preprint arXiv:2307.13269}
}

@article{zadouri2023pushingmoe,
      title={Pushing Mixture of Experts to the Limit: Extremely Parameter Efficient MoE for Instruction Tuning}, 
      author={Ted Zadouri and Ahmet Üstün and Arash Ahmadian and Beyza Ermiş and Acyr Locatelli and Sara Hooker},
      year={2023},
      journal={arXiv preprint arXiv:2309.05444},
      archivePrefix={arXiv},
      primaryClass={cs.CL},
}

@inproceedings{feng2024moelora,
    title = "Mixture-of-{L}o{RA}s: An Efficient Multitask Tuning Method for Large Language Models",
    author = "Feng, Wenfeng  and
      Hao, Chuzhan  and
      Zhang, Yuewei  and
      Han, Yu  and
      Wang, Hao",
    booktitle = "Proceedings of the 2024 Joint International Conference on Computational Linguistics, Language Resources and Evaluation (LREC-COLING 2024)",
    month = may,
    year = "2024",
    address = "Torino, Italia",
    publisher = "ELRA and ICCL",
    pages = "11371--11380",
}

@article{yang2024moral,
      title={MoRAL: MoE Augmented LoRA for LLMs' Lifelong Learning}, 
      author={Shu Yang and Muhammad Asif Ali and Cheng-Long Wang and Lijie Hu and Di Wang},
      year={2024},
      journal={arXiv preprint arXiv:2402.11260},
      archivePrefix={arXiv},
      primaryClass={cs.CL},
}

@inproceedings{dou2024loramoe,
    title = "{L}o{RAM}o{E}: Alleviating World Knowledge Forgetting in Large Language Models via {M}o{E}-Style Plugin",
    author = "Dou, Shihan  and
      Zhou, Enyu  and
      Liu, Yan  and
      Gao, Songyang  and
      Shen, Wei  and
      Xiong, Limao  and
      Zhou, Yuhao  and
      Wang, Xiao  and
      Xi, Zhiheng  and
      Fan, Xiaoran  and
      Pu, Shiliang  and
      Zhu, Jiang  and
      Zheng, Rui  and
      Gui, Tao  and
      Zhang, Qi  and
      Huang, Xuanjing",
    booktitle = "Proceedings of the 62nd Annual Meeting of the Association for Computational Linguistics (Volume 1: Long Papers)",
    month = aug,
    year = "2024",
    address = "Bangkok, Thailand",
    publisher = "Association for Computational Linguistics",
    pages = "1932--1945",
}

@article{gou2024loramoevl,
      title={Mixture of Cluster-conditional LoRA Experts for Vision-language Instruction Tuning}, 
      author={Yunhao Gou and Zhili Liu and Kai Chen and Lanqing Hong and Hang Xu and Aoxue Li and Dit-Yan Yeung and James T. Kwok and Yu Zhang},
      year={2024},
      journal={arXiv preprint arXiv:2312.12379},
      archivePrefix={arXiv},
      primaryClass={cs.CV},
}

@article{liu2024loramoemed,
      title={When MOE Meets LLMs: Parameter Efficient Fine-tuning for Multi-task Medical Applications}, 
      author={Qidong Liu and Xian Wu and Xiangyu Zhao and Yuanshao Zhu and Derong Xu and Feng Tian and Yefeng Zheng},
      year={2024},
      journal={arXiv preprint arXiv:2310.18339},
      archivePrefix={arXiv},
      primaryClass={cs.CL},
}

@inproceedings{rajbhandari2022deepspeed-moe,
author = {Rajbhandari, Samyam and Li, Conglong and Yao, Zhewei and Zhang, Minjia and Yazdani Aminabadi, Reza and Awan, Ammar Ahmad and Rasley, Jeff and He, Yuxiong},
title = {DeepSpeed-MoE: Advancing Mixture-of-Experts Inference and Training to Power Next-Generation AI Scale},
booktitle = {ICML 2022},
year = {2022},
month = {January},
}

@InProceedings{pmlr-v162-clark22a,
  title = 	 {Unified Scaling Laws for Routed Language Models},
  author =       {Clark, Aidan and De Las Casas, Diego and Guy, Aurelia and Mensch, Arthur and Paganini, Michela and Hoffmann, Jordan and Damoc, Bogdan and Hechtman, Blake and Cai, Trevor and Borgeaud, Sebastian and Van Den Driessche, George Bm and Rutherford, Eliza and Hennigan, Tom and Johnson, Matthew J and Cassirer, Albin and Jones, Chris and Buchatskaya, Elena and Budden, David and Sifre, Laurent and Osindero, Simon and Vinyals, Oriol and Ranzato, Marc'Aurelio and Rae, Jack and Elsen, Erich and Kavukcuoglu, Koray and Simonyan, Karen},
  booktitle = 	 {Proceedings of the 39th International Conference on Machine Learning},
  pages = 	 {4057--4086},
  year = 	 {2022},
  volume = 	 {162},
  series = 	 {Proceedings of Machine Learning Research},
  month = 	 {17--23 Jul},
  publisher =    {PMLR},
}

@INPROCEEDINGS {he2015ImageNet,
author = { He, Kaiming and Zhang, Xiangyu and Ren, Shaoqing and Sun, Jian },
booktitle = { 2015 IEEE International Conference on Computer Vision (ICCV) },
title = {{ Delving Deep into Rectifiers: Surpassing Human-Level Performance on ImageNet Classification }},
year = {2015},
volume = {},
ISSN = {2380-7504},
pages = {1026-1034},
publisher = {IEEE Computer Society},
address = {Los Alamitos, CA, USA},
month =Dec}

@article{cilia2022diagnosing,
  title={Diagnosing Alzheimer’s disease from on-line handwriting: A novel dataset and performance benchmarking},
  author={Cilia, Nicole D and De Gregorio, Giuseppe and De Stefano, Claudio and Fontanella, Francesco and Marcelli, Angelo and Parziale, Antonio},
  journal={Engineering Applications of Artificial Intelligence},
  volume={111},
  pages={104822},
  year={2022},
  publisher={Elsevier}
}

\newpage 

\appendix

\section{Results of Classification by Task}
This appendix provides detailed results for the task-level classification experiments described in the main paper. The tables report the performance of MLP, MoE, and LoRA-MoE architectures across the 25 handwriting tasks of the DARWIN dataset under different model configurations. For each experiment, the mean results obtained over repeated runs are presented using several evaluation metrics, including accuracy, sensitivity, and specificity. The configurations correspond to different hidden dimensions (5–25) as well as stacking strategies (Max and Mean). These tables complement the results reported in the main text by providing a detailed view of the performance of each model across individual handwriting tasks.

\setlength{\tabcolsep}{3pt}   % default is 6pt
\renewcommand{\arraystretch}{0.9}
\begin{table}[h!]
\centering
\caption{Accuracy (\%) of MLP configurations on each experiment.}
\label{tab:mlp_accuracy}
\small
\setlength{\tabcolsep}{3pt}
\renewcommand{\arraystretch}{0.9}
\begin{tabular}{lccccccc}

\toprule
\textbf{Exp} & \textbf{MLP-5} & \textbf{MLP-10} & \textbf{MLP-15} & \textbf{MLP-20} & \textbf{MLP-25} & \textbf{MLP-Max} & \textbf{MLP-Mean} \\
\midrule
1 & 57.57 & 61.86 & 63.86 & 63.29 & 63.00 & 63.14 & 64.14 \\
2 & 64.29 & 68.00 & 67.14 & 69.14 & 70.14 & 69.57 & 69.14 \\
3 & 65.71 & 67.43 & 68.57 & 67.29 & 68.71 & 68.00 & 67.43 \\
4 & 62.43 & 66.43 & 66.29 & 66.29 & 67.71 & 66.86 & 66.43 \\
5 & 64.14 & 72.00 & 72.14 & 70.71 & 70.00 & 71.86 & 72.14 \\
6 & 65.14 & 71.43 & 70.43 & 72.14 & 71.43 & 71.43 & 71.43 \\
7 & 65.00 & 74.43 & 73.43 & 74.57 & 74.14 & 74.14 & 75.00 \\
8 & 63.57 & 73.43 & 74.29 & 73.43 & 72.57 & 73.71 & 73.71 \\
9 & 70.29 & 74.00 & 75.71 & 74.71 & 74.43 & 74.71 & 74.71 \\
10 & 62.57 & 70.00 & 67.71 & 69.00 & 70.43 & 69.57 & 69.14 \\
11 & 63.86 & 67.43 & 67.00 & 66.86 & 67.71 & 67.00 & 66.71 \\
12 & 57.86 & 67.57 & 70.14 & 68.57 & 69.43 & 69.14 & 69.29 \\
13 & 62.71 & 68.00 & 68.29 & 68.00 & 69.29 & 68.14 & 69.57 \\
14 & 60.43 & 65.14 & 68.29 & 68.00 & 70.71 & 69.71 & 69.29 \\
15 & 70.29 & 72.43 & 73.00 & 73.14 & 73.29 & 74.57 & 74.14 \\
16 & 63.86 & 72.57 & 75.14 & 74.71 & 77.57 & 77.29 & 76.57 \\
17 & 65.86 & 75.14 & 76.71 & 79.14 & 79.00 & 79.00 & 79.29 \\
18 & 64.71 & 65.86 & 68.57 & 71.00 & 70.86 & 69.86 & 69.71 \\
19 & 63.57 & 72.14 & 73.71 & 73.86 & 75.14 & 75.43 & 76.29 \\
20 & 61.57 & 69.29 & 70.57 & 70.43 & 71.71 & 70.14 & 70.43 \\
21 & 63.29 & 65.29 & 66.86 & 70.29 & 71.00 & 67.86 & 67.43 \\
22 & 62.86 & 67.29 & 69.00 & 71.00 & 70.71 & 70.00 & 69.57 \\
23 & 67.57 & 75.43 & 75.00 & 76.00 & 74.43 & 75.29 & 75.57 \\
24 & 63.43 & 71.71 & 75.14 & 76.29 & 74.71 & 76.71 & 76.57 \\
25 & 68.86 & 70.43 & 71.43 & 70.43 & 70.86 & 72.00 & 72.86 \\
\bottomrule
\end{tabular}
\end{table}

\begin{table}[h!]
\centering
\caption{Accuracy (\%) of MoE configurations on each experiment.}
\label{tab:moe_accuracy}
\small
\begin{tabular}{lccccccc}
\toprule
\textbf{Exp} & \textbf{MoE-5} & \textbf{MoE-10} & \textbf{MoE-15} & \textbf{MoE-20} & \textbf{MoE-25} & \textbf{MoE-Max} & \textbf{MoE-Mean} \\
\midrule
1 & 57.57 & 61.57 & 60.71 & 61.00 & 62.57 & 60.29 & 60.29 \\
2 & 60.57 & 65.14 & 66.57 & 68.57 & 67.14 & 68.71 & 68.71 \\
3 & 59.71 & 65.29 & 66.14 & 66.71 & 67.00 & 68.86 & 68.86 \\
4 & 60.29 & 65.71 & 65.00 & 63.86 & 65.43 & 65.14 & 65.14 \\
5 & 66.14 & 62.71 & 67.43 & 69.57 & 68.00 & 70.86 & 70.86 \\
6 & 60.29 & 66.43 & 69.00 & 69.43 & 68.00 & 72.43 & 72.43 \\
7 & 66.86 & 72.71 & 71.14 & 73.14 & 72.43 & 74.29 & 74.29 \\
8 & 61.86 & 69.00 & 69.57 & 70.14 & 70.43 & 72.86 & 72.86 \\
9 & 63.71 & 70.57 & 72.71 & 72.71 & 73.29 & 75.57 & 75.57 \\
10 & 61.43 & 65.14 & 66.57 & 68.14 & 66.57 & 69.14 & 69.14 \\
11 & 60.43 & 67.71 & 65.57 & 67.29 & 68.86 & 68.86 & 67.43 \\
12 & 61.14 & 66.14 & 65.29 & 66.14 & 70.00 & 70.57 & 70.57 \\
13 & 61.00 & 64.57 & 65.86 & 66.29 & 65.86 & 68.57 & 68.57 \\
14 & 57.57 & 64.00 & 66.71 & 69.43 & 67.71 & 70.29 & 70.29 \\
15 & 59.43 & 69.86 & 67.43 & 72.86 & 68.14 & 71.71 & 71.71 \\
16 & 62.14 & 69.57 & 71.86 & 70.71 & 72.29 & 74.43 & 74.43 \\
17 & 67.86 & 69.86 & 73.86 & 77.86 & 77.29 & 79.00 & 78.00 \\
18 & 61.57 & 62.43 & 64.29 & 66.57 & 67.57 & 69.29 & 69.29 \\
19 & 63.29 & 66.71 & 69.00 & 71.00 & 71.71 & 73.57 & 72.43 \\
20 & 64.00 & 68.00 & 67.29 & 71.43 & 68.86 & 72.71 & 72.71 \\
21 & 58.00 & 67.14 & 65.71 & 72.00 & 70.14 & 69.57 & 69.57 \\
22 & 59.43 & 66.86 & 67.14 & 68.00 & 69.71 & 70.14 & 70.14 \\
23 & 66.86 & 69.86 & 70.86 & 75.00 & 74.71 & 75.86 & 77.14 \\
24 & 60.14 & 67.43 & 69.71 & 71.79 & 68.86 & 72.43 & 74.57 \\
25 & 63.00 & 67.00 & 67.00 & 69.71 & 68.71 & 71.86 & 72.14 \\
\bottomrule
\end{tabular}
\end{table}

\begin{table}[htbp]
\centering
\caption{Sensitivity of LM (LoRA-MoE) configurations on each experiment.}
\label{tab:lm_sensitivity}
\small
\begin{tabular}{lccccccc}
\toprule
\textbf{Exp} & \textbf{LM-5} & \textbf{LM-10} & \textbf{LM-15} & \textbf{LM-20} & \textbf{LM-25} & \textbf{LM-Max} & \textbf{LM-Mean} \\
\midrule
1 & 60.83 & 60.83 & 59.17 & 57.22 & 57.78 & 56.94 & 56.94 \\
2 & 72.78 & 63.06 & 66.67 & 70.56 & 69.72 & 69.72 & 68.89 \\
3 & 68.61 & 66.67 & 67.22 & 69.72 & 67.78 & 69.72 & 68.06 \\
4 & 67.22 & 66.94 & 62.22 & 66.67 & 67.50 & 67.50 & 66.94 \\
5 & 64.44 & 62.22 & 63.89 & 63.61 & 62.22 & 64.17 & 63.61 \\
6 & 68.61 & 59.17 & 60.28 & 64.44 & 61.67 & 63.89 & 62.78 \\
7 & 73.89 & 72.78 & 73.89 & 69.72 & 71.67 & 73.33 & 73.06 \\
8 & 66.94 & 62.22 & 64.17 & 66.11 & 67.22 & 64.72 & 65.00 \\
9 & 59.44 & 65.56 & 69.72 & 69.72 & 70.28 & 68.06 & 66.11 \\
10 & 72.78 & 63.06 & 66.67 & 70.56 & 69.72 & 69.72 & 68.89 \\
11 & 68.61 & 66.67 & 67.22 & 69.72 & 67.78 & 69.72 & 68.06 \\
12 & 67.22 & 66.94 & 62.22 & 66.67 & 67.50 & 67.50 & 66.94 \\
13 & 64.72 & 60.28 & 57.22 & 60.56 & 60.00 & 59.17 & 59.44 \\
14 & 74.17 & 68.06 & 67.78 & 72.22 & 69.72 & 73.06 & 73.89 \\
15 & 54.44 & 67.78 & 66.11 & 69.17 & 68.61 & 69.44 & 68.89 \\
16 & 67.22 & 70.56 & 71.39 & 73.61 & 75.83 & 74.44 & 73.33 \\
17 & 75.00 & 76.94 & 79.17 & 76.94 & 76.67 & 79.72 & 78.06 \\
18 & 71.11 & 69.44 & 62.50 & 69.72 & 69.44 & 69.72 & 68.33 \\
19 & 63.33 & 72.50 & 71.11 & 69.44 & 70.28 & 71.11 & 71.94 \\
20 & 70.56 & 64.72 & 67.78 & 67.78 & 70.28 & 69.17 & 67.78 \\
21 & 72.22 & 69.72 & 71.67 & 70.28 & 71.94 & 69.72 & 70.00 \\
22 & 69.72 & 68.06 & 68.06 & 67.78 & 69.44 & 69.44 & 69.17 \\
23 & 70.56 & 75.28 & 74.17 & 73.61 & 73.61 & 72.78 & 72.78 \\
24 & 67.22 & 70.56 & 70.00 & 68.33 & 68.89 & 71.67 & 71.67 \\
25 & 74.17 & 73.06 & 70.00 & 71.39 & 70.00 & 73.06 & 73.61 \\
\bottomrule
\end{tabular}
\end{table}

\begin{table}[htbp]
\centering
\caption{Sensitivity of MLP configurations on each experiment.}
\label{tab:mlp_sensitivity}
\small
\begin{tabular}{lccccccc}
\toprule
\textbf{Exp} & \textbf{MLP-5} & \textbf{MLP-10} & \textbf{MLP-15} & \textbf{MLP-20} & \textbf{MLP-25} & \textbf{MLP-Max} & \textbf{MLP-Mean} \\
\midrule
1 & 50.00 & 59.72 & 60.28 & 58.33 & 56.94 & 57.22 & 57.50 \\
2 & 70.00 & 65.00 & 64.44 & 66.11 & 67.78 & 67.50 & 65.83 \\
3 & 63.89 & 68.61 & 69.44 & 68.61 & 68.33 & 70.00 & 69.72 \\
4 & 63.89 & 65.56 & 65.83 & 65.00 & 65.28 & 66.39 & 66.67 \\
5 & 59.72 & 60.83 & 60.56 & 60.83 & 60.28 & 60.56 & 60.83 \\
6 & 60.28 & 60.28 & 62.22 & 63.06 & 64.72 & 63.06 & 62.50 \\
7 & 79.17 & 75.00 & 72.22 & 73.33 & 73.06 & 72.78 & 73.33 \\
8 & 64.44 & 64.17 & 63.06 & 64.17 & 63.61 & 64.44 & 64.17 \\
9 & 63.06 & 64.44 & 66.39 & 66.67 & 68.33 & 66.39 & 66.11 \\
10 & 67.50 & 65.00 & 64.44 & 66.11 & 67.78 & 67.50 & 65.83 \\
11 & 76.39 & 68.61 & 69.44 & 68.61 & 68.33 & 70.00 & 69.72 \\
12 & 63.89 & 65.56 & 65.83 & 65.00 & 65.28 & 66.39 & 66.67 \\
13 & 69.44 & 63.61 & 58.89 & 59.72 & 60.56 & 58.89 & 60.00 \\
14 & 66.39 & 71.11 & 72.22 & 72.22 & 72.78 & 75.28 & 75.28 \\
15 & 71.39 & 69.72 & 68.06 & 67.78 & 68.61 & 69.72 & 68.61 \\
16 & 64.17 & 70.28 & 70.56 & 70.28 & 73.61 & 73.33 & 72.22 \\
17 & 58.33 & 76.67 & 76.39 & 75.56 & 77.78 & 77.50 & 76.94 \\
18 & 68.33 & 63.61 & 62.78 & 67.78 & 67.50 & 67.50 & 66.11 \\
19 & 68.61 & 73.06 & 72.50 & 71.39 & 71.39 & 73.06 & 73.61 \\
20 & 67.78 & 68.89 & 67.50 & 69.44 & 69.44 & 68.06 & 68.33 \\
21 & 58.33 & 68.89 & 70.28 & 71.11 & 71.94 & 69.44 & 71.11 \\
22 & 60.28 & 71.11 & 66.39 & 67.78 & 67.78 & 69.44 & 68.89 \\
23 & 57.78 & 71.94 & 72.22 & 73.89 & 71.67 & 71.94 & 72.50 \\
24 & 73.61 & 74.17 & 70.28 & 71.94 & 71.94 & 73.89 & 74.17 \\
25 & 73.89 & 73.61 & 73.61 & 70.56 & 71.39 & 72.22 & 74.17 \\
\bottomrule
\end{tabular}
\end{table}

\begin{table}[htbp]
\centering
\caption{Sensitivity of MoE configurations on each experiment.}
\label{tab:moe_sensitivity}
\small
\begin{tabular}{lccccccc}
\toprule
\textbf{Exp} & \textbf{MoE-5} & \textbf{MoE-10} & \textbf{MoE-15} & \textbf{MoE-20} & \textbf{MoE-25} & \textbf{MoE-Max} & \textbf{MoE-Mean} \\
\midrule
1 & 56.11 & 58.06 & 56.94 & 61.39 & 61.11 & 58.33 & 58.06 \\
2 & 65.00 & 61.94 & 65.83 & 64.72 & 63.89 & 68.06 & 66.94 \\
3 & 58.89 & 60.28 & 58.33 & 60.00 & 62.50 & 65.83 & 65.00 \\
4 & 57.22 & 60.83 & 58.06 & 56.67 & 59.17 & 58.06 & 57.78 \\
5 & 62.50 & 61.39 & 62.50 & 60.28 & 63.06 & 65.28 & 64.44 \\
6 & 54.17 & 57.78 & 61.39 & 64.44 & 62.50 & 65.83 & 64.44 \\
7 & 67.78 & 75.00 & 71.39 & 72.50 & 72.50 & 74.17 & 74.44 \\
8 & 55.00 & 61.94 & 65.83 & 64.17 & 63.06 & 65.28 & 63.61 \\
9 & 61.94 & 65.83 & 74.44 & 70.83 & 71.94 & 72.78 & 70.83 \\
10 & 63.33 & 62.50 & 65.83 & 68.33 & 65.28 & 68.06 & 67.78 \\
11 & 61.94 & 72.22 & 69.72 & 71.67 & 71.67 & 73.61 & 74.72 \\
12 & 61.11 & 63.89 & 64.72 & 63.89 & 69.44 & 71.67 & 70.28 \\
13 & 59.17 & 58.61 & 60.83 & 61.11 & 63.33 & 64.44 & 63.61 \\
14 & 62.78 & 66.67 & 68.89 & 71.94 & 70.28 & 73.06 & 73.06 \\
15 & 53.89 & 69.44 & 68.89 & 70.00 & 63.33 & 68.06 & 69.44 \\
16 & 62.50 & 69.17 & 70.28 & 68.61 & 72.50 & 72.50 & 73.06 \\
17 & 68.33 & 73.06 & 75.56 & 80.00 & 79.44 & 80.00 & 79.44 \\
18 & 68.33 & 59.72 & 62.22 & 67.22 & 64.72 & 68.89 & 66.39 \\
19 & 68.89 & 68.61 & 68.61 & 70.00 & 70.83 & 71.11 & 68.89 \\
20 & 62.50 & 66.11 & 66.94 & 71.67 & 70.00 & 70.83 & 71.11 \\
21 & 63.61 & 67.50 & 67.78 & 71.39 & 70.83 & 71.39 & 70.83 \\
22 & 60.00 & 72.50 & 67.78 & 69.17 & 68.61 & 69.44 & 70.56 \\
23 & 69.44 & 70.83 & 69.44 & 76.39 & 75.28 & 75.28 & 76.94 \\
24 & 60.56 & 65.83 & 67.78 & 69.44 & 67.22 & 70.56 & 72.50 \\
25 & 65.00 & 68.89 & 69.44 & 69.44 & 68.06 & 72.78 & 72.50 \\
\bottomrule
\end{tabular}
\end{table}

\begin{table}[h!]
\centering
\caption{Specificity of LM (LoRA-MoE) configurations on each experiment.}
\label{tab:lm_specificity}
\small
\begin{tabular}{lccccccc}
\toprule
\textbf{Exp} & \textbf{LM-5} & \textbf{LM-10} & \textbf{LM-15} & \textbf{LM-20} & \textbf{LM-25} & \textbf{LM-Max} & \textbf{LM-Mean} \\
\midrule
1 & 60.00 & 63.53 & 69.12 & 67.65 & 64.71 & 66.76 & 70.88 \\
2 & 63.24 & 75.00 & 71.18 & 70.59 & 70.88 & 72.35 & 72.94 \\
3 & 65.29 & 65.29 & 66.47 & 66.18 & 69.41 & 67.94 & 67.35 \\
4 & 58.24 & 67.94 & 74.12 & 70.29 & 75.29 & 73.53 & 73.53 \\
5 & 64.12 & 74.41 & 76.47 & 73.53 & 74.12 & 75.88 & 76.47 \\
6 & 66.47 & 80.00 & 78.53 & 75.29 & 74.41 & 78.24 & 78.82 \\
7 & 74.41 & 75.59 & 72.35 & 78.82 & 77.35 & 77.94 & 78.24 \\
8 & 74.12 & 82.06 & 82.35 & 82.94 & 79.41 & 82.65 & 83.82 \\
9 & 83.24 & 83.82 & 81.76 & 77.94 & 78.53 & 81.47 & 84.41 \\
10 & 63.24 & 75.00 & 71.18 & 70.59 & 70.88 & 72.35 & 72.94 \\
11 & 65.29 & 65.29 & 66.47 & 66.18 & 69.41 & 67.94 & 67.35 \\
12 & 58.24 & 67.94 & 74.12 & 70.29 & 75.29 & 73.53 & 73.53 \\
13 & 64.12 & 74.41 & 76.47 & 73.53 & 74.12 & 75.88 & 76.47 \\
14 & 48.53 & 60.29 & 68.82 & 67.06 & 69.71 & 67.35 & 66.47 \\
15 & 76.47 & 77.94 & 77.65 & 77.06 & 77.94 & 77.35 & 78.82 \\
16 & 65.29 & 77.65 & 77.94 & 75.29 & 74.71 & 76.47 & 77.94 \\
17 & 71.76 & 78.24 & 76.18 & 75.59 & 77.35 & 80.88 & 81.76 \\
18 & 53.53 & 67.06 & 74.71 & 71.18 & 72.06 & 70.88 & 71.47 \\
19 & 72.94 & 71.18 & 79.41 & 75.29 & 78.82 & 78.24 & 80.00 \\
20 & 61.47 & 72.94 & 70.88 & 72.65 & 70.00 & 72.65 & 72.94 \\
21 & 59.12 & 69.41 & 65.88 & 68.82 & 73.53 & 69.12 & 68.53 \\
22 & 64.12 & 73.82 & 72.65 & 72.65 & 72.35 & 71.47 & 71.47 \\
23 & 72.35 & 75.29 & 77.94 & 73.82 & 75.00 & 77.06 & 77.65 \\
24 & 60.00 & 71.18 & 78.53 & 75.88 & 77.06 & 77.94 & 79.41 \\
25 & 61.18 & 68.82 & 72.35 & 70.88 & 69.12 & 71.76 & 72.35 \\
\bottomrule
\end{tabular}
\end{table}

\begin{table}[htbp]
\centering
\caption{Specificity of MLP configurations on each experiment.}
\label{tab:mlp_specificity}
\small
\begin{tabular}{lccccccc}
\toprule
\textbf{Exp} & \textbf{MLP-5} & \textbf{MLP-10} & \textbf{MLP-15} & \textbf{MLP-20} & \textbf{MLP-25} & \textbf{MLP-Max} & \textbf{MLP-Mean} \\
\midrule
1 & 65.59 & 64.12 & 67.65 & 68.53 & 69.41 & 69.41 & 71.18 \\
2 & 62.35 & 75.29 & 71.76 & 72.06 & 73.24 & 71.76 & 72.65 \\
3 & 67.94 & 72.65 & 74.71 & 74.41 & 75.59 & 73.53 & 74.12 \\
4 & 51.47 & 72.06 & 74.71 & 77.65 & 77.24 & 72.06 & 72.06 \\
5 & 55.59 & 72.65 & 78.24 & 76.76 & 78.53 & 77.94 & 79.41 \\
6 & 70.29 & 83.24 & 79.12 & 81.76 & 78.53 & 80.29 & 80.88 \\
7 & 50.00 & 73.82 & 74.71 & 75.88 & 75.29 & 75.59 & 76.76 \\
8 & 62.65 & 83.24 & 86.18 & 83.24 & 82.06 & 83.53 & 83.82 \\
9 & 77.94 & 84.12 & 85.59 & 83.24 & 80.88 & 83.53 & 83.82 \\
10 & 62.35 & 75.29 & 71.76 & 72.06 & 73.24 & 71.76 & 72.65 \\
11 & 50.59 & 66.18 & 64.41 & 65.00 & 67.06 & 63.82 & 63.53 \\
12 & 51.47 & 69.71 & 74.71 & 72.35 & 73.82 & 72.06 & 72.06 \\
13 & 55.59 & 72.65 & 78.24 & 76.76 & 78.53 & 77.94 & 79.71 \\
14 & 54.12 & 58.82 & 64.12 & 63.53 & 68.53 & 63.82 & 62.94 \\
15 & 69.12 & 75.29 & 78.24 & 78.82 & 78.24 & 79.71 & 80.00 \\
16 & 63.53 & 75.00 & 80.00 & 79.41 & 81.76 & 81.47 & 81.18 \\
17 & 73.82 & 73.53 & 77.06 & 82.94 & 80.29 & 80.59 & 81.76 \\
18 & 60.88 & 68.24 & 74.71 & 74.41 & 74.41 & 72.35 & 73.53 \\
19 & 58.24 & 71.18 & 75.00 & 76.47 & 79.12 & 77.94 & 79.12 \\
20 & 55.00 & 69.71 & 73.82 & 71.47 & 74.12 & 72.35 & 72.65 \\
21 & 68.53 & 61.47 & 63.24 & 69.41 & 70.00 & 66.18 & 63.53 \\
22 & 65.59 & 63.24 & 71.76 & 74.41 & 73.82 & 70.59 & 70.29 \\
23 & 77.94 & 79.12 & 77.94 & 78.24 & 77.35 & 78.82 & 78.82 \\
24 & 52.65 & 69.12 & 80.29 & 80.88 & 77.65 & 79.71 & 79.12 \\
25 & 63.53 & 67.06 & 69.12 & 70.29 & 70.29 & 71.76 & 71.47 \\
\bottomrule
\end{tabular}
\end{table}

\begin{table}[htbp]
\centering
\caption{Specificity of MoE configurations on each experiment.}
\label{tab:moe_specificity}
\small
\begin{tabular}{lccccccc}
\toprule
\textbf{Exp} & \textbf{MoE-5} & \textbf{MoE-10} & \textbf{MoE-15} & \textbf{MoE-20} & \textbf{MoE-25} & \textbf{MoE-Max} & \textbf{MoE-Mean} \\
\midrule
1 & 59.12 & 64.71 & 60.59 & 64.12 & 61.18 & 62.35 & 65.29 \\
2 & 55.88 & 68.53 & 67.35 & 72.65 & 70.59 & 69.41 & 72.35 \\
3 & 60.59 & 70.59 & 69.12 & 71.76 & 71.76 & 72.06 & 72.06 \\
4 & 63.53 & 70.88 & 72.35 & 71.47 & 72.06 & 72.65 & 75.88 \\
5 & 70.00 & 64.12 & 72.65 & 79.41 & 73.24 & 76.76 & 78.24 \\
6 & 66.76 & 75.59 & 77.06 & 74.71 & 73.82 & 79.41 & 80.88 \\
7 & 65.88 & 70.29 & 70.88 & 73.82 & 72.35 & 74.41 & 76.18 \\
8 & 69.12 & 76.47 & 73.53 & 76.47 & 78.24 & 80.88 & 79.41 \\
9 & 65.59 & 75.59 & 70.88 & 74.71 & 74.71 & 78.53 & 78.24 \\
10 & 59.41 & 67.94 & 67.35 & 67.94 & 67.94 & 70.29 & 71.47 \\
11 & 58.82 & 62.94 & 61.18 & 62.65 & 65.88 & 60.88 & 62.65 \\
12 & 61.18 & 68.53 & 65.88 & 68.53 & 70.59 & 69.41 & 71.76 \\
13 & 62.94 & 70.88 & 71.18 & 71.76 & 68.53 & 72.94 & 74.41 \\
14 & 52.06 & 61.18 & 64.41 & 66.76 & 65.00 & 67.35 & 67.06 \\
15 & 65.29 & 70.29 & 65.88 & 75.88 & 73.24 & 75.59 & 77.06 \\
16 & 61.76 & 70.00 & 73.53 & 72.94 & 72.06 & 76.47 & 77.94 \\
17 & 67.35 & 66.47 & 72.06 & 75.59 & 75.00 & 77.94 & 77.35 \\
18 & 54.41 & 65.29 & 66.47 & 65.88 & 70.59 & 69.71 & 72.35 \\
19 & 57.35 & 64.71 & 69.41 & 72.06 & 72.65 & 76.18 & 76.18 \\
20 & 65.59 & 70.00 & 67.65 & 71.18 & 67.65 & 74.71 & 75.00 \\
21 & 52.06 & 66.76 & 63.53 & 72.65 & 69.41 & 67.65 & 68.82 \\
22 & 58.82 & 60.88 & 66.47 & 66.76 & 70.88 & 70.88 & 70.00 \\
23 & 64.12 & 68.82 & 72.35 & 73.53 & 74.12 & 76.47 & 77.35 \\
24 & 59.71 & 69.12 & 71.76 & 74.12 & 70.59 & 74.41 & 76.76 \\
25 & 60.88 & 65.00 & 64.41 & 70.00 & 69.41 & 70.88 & 71.76 \\
\bottomrule
\end{tabular}
\end{table}

\end{document}